\documentclass[10pt,twocolumn,letterpaper]{article}

\usepackage{iccv}
\usepackage{times}
\usepackage{epsfig}
\usepackage{graphicx}
\usepackage{amsmath}
\usepackage{amssymb}

\newcommand{\nn}{\mathrm{K}}
\newcommand{\cR}{R^{c}}
\newcommand{\cT}{t^{c}}
\newcommand{\depth}{\mathrm{d}}
\newcommand{\masks}{\mathrm{m}}

\begin{document}
\iccvfinalcopy
\title{SfM-Net: Learning of Structure and Motion from Video}

\author{Sudheendra Vijayanarasimhan\thanks{Google Research}\\
{\tt\small svnaras@google.com}
\and
Susanna Ricco\footnotemark[1]\\
{\tt\small ricco@google.com}
\and
Cordelia Schmid\thanks{Inria, Grenoble, France}\textsuperscript{  }\footnotemark[1] \\
{\tt\small cordelia.schmid@inria.fr}
\and
Rahul Sukthankar\footnotemark[1] \\
{\tt\small sukthankar@google.com}
\and
Katerina Fragkiadaki\thanks{Carnegie Mellon University} \\
{\tt\small katef@cs.cmu.edu}
}

\maketitle

\begin{abstract}
  We propose SfM-Net, a geometry-aware neural network for motion estimation in videos that decomposes frame-to-frame pixel motion in terms of scene and object depth, camera motion and 3D object rotations and translations. Given a  sequence of frames, SfM-Net predicts depth, segmentation, camera and rigid object motions, converts those into a dense  frame-to-frame motion  field (optical flow), differentiably warps frames in time to match pixels and back-propagates. The model can be trained with various degrees of supervision: 1) self-supervised by the re-projection photometric error (completely unsupervised), 2) supervised by ego-motion (camera motion), or 3) supervised by depth (e.g., as provided by RGBD sensors). SfM-Net  extracts meaningful depth estimates 
and successfully estimates frame-to-frame camera rotations and translations. It often  successfully segments the moving objects in the scene, even though such supervision is never provided.  

\end{abstract}

\section{Introduction}
\label{sec:introduction}

\begin{figure}
\centering
\includegraphics[width=.95\linewidth]{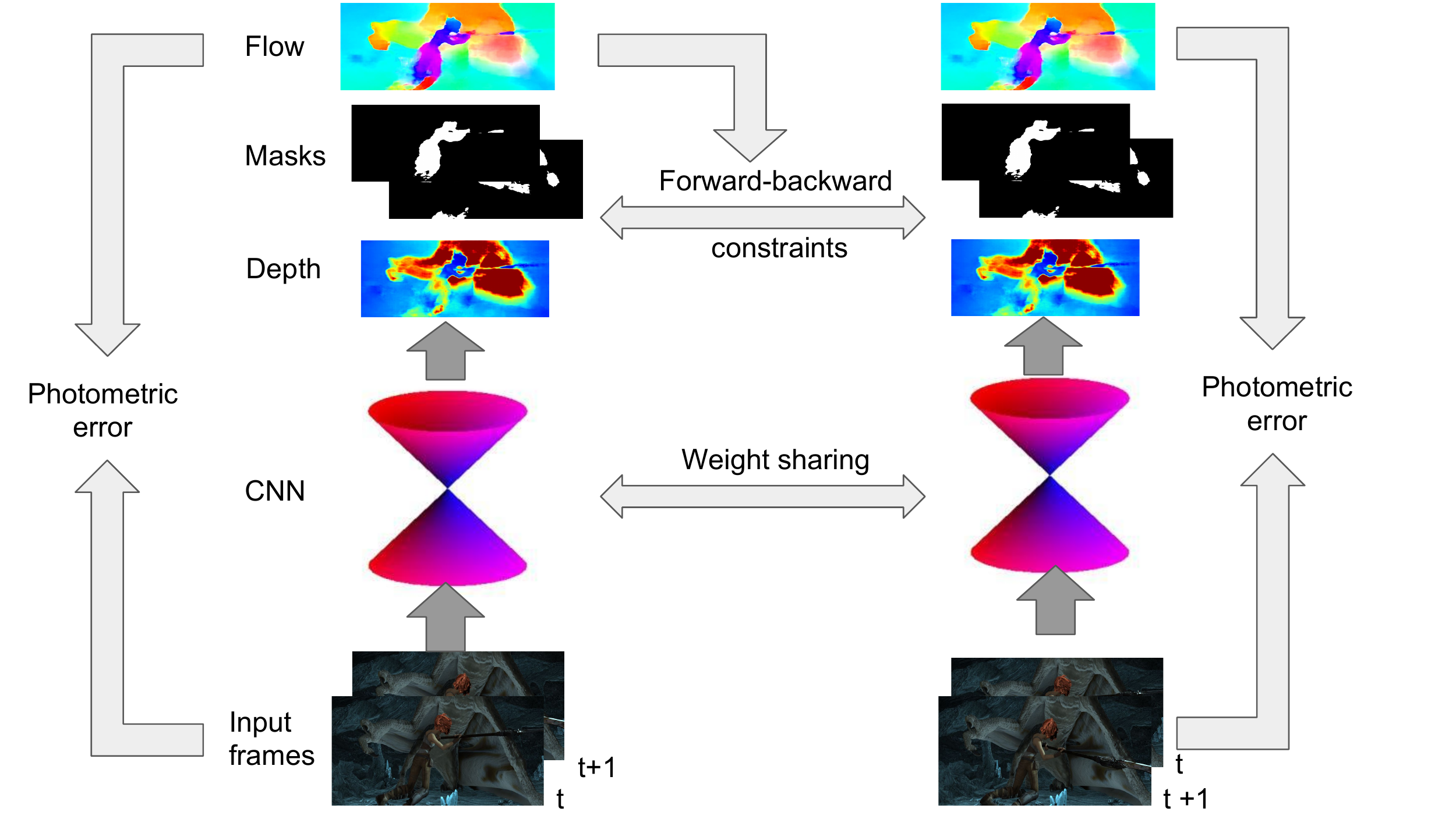}
\caption{SfM-Net: Given a pair of frames as input, our model decomposes  frame-to-frame pixel motion into 3D scene depth, 3D camera rotation and translation, a set of motion masks and corresponding 3D rigid rotations and translations. It  backprojects the resulting 3D scene flow into 2D optical flow and warps accordingly to match pixels from one frame to the next. Forward-backward consistency checks constrain the estimated depth.  
}
\label{fig:SfMNet}
\end{figure}

We propose SfM-Net, a neural network that is trained to extract 3D structure,  ego-motion, segmentation, object rotations and translations in an end-to-end fashion in videos,  by exploiting  the geometry of image formation. 
Given a pair of frames and camera intrinsics, SfM-Net, depicted in Figure~\ref{fig:SfMNet},   computes  depth,  3D camera motion,  a set of 3D rotations and translations for the dynamic objects in the scene, and corresponding pixel assignment masks. 
Those in turn provide a geometrically meaningful motion field (optical flow) that is used to differentiably warp each frame to the next. Pixel matching across consecutive frames, constrained by forward-backward consistency on the computed motion and 3D structure, provides gradients during training in the case of self-supervision.  
SfM-Net can take advantage of varying levels of supervision, as demonstrated in our experiments: completely unsupervised (self-supervised), supervised by camera motion, or supervised by depth (from Kinect).

SfM-Net is inspired by works that impose geometric constraints on optical flow,  exploiting rigidity of the visual scene, such as early low-parametric optical flow methods~\cite{Bergen:1992:HMM:645305.648720,citeulike:2638091,Nir_IJCV2008} or the so-called direct methods for visual SLAM (Simultaneous Localization and Mapping) that perform dense pixel matching from frame to frame while estimating a camera trajectory and depth of the pixels in the scene~\cite{engel14eccv,schoeps14ismar}. In contrast to those, instead of optimizing directly over optical flow vectors, 3D point coordinates or camera rotation and translation, our model optimizes over neural network weights that, given a pair of frames, produce such 3D structure and motion. In this way, our method learns to estimate structure and motion, and can in principle improve as it processes more videos, in contrast to non-learning based alternatives. It can thus be made robust to lack of texture, degenerate camera motion trajectories or dynamic objects (our model explicitly accounts for those), by providing appropriate supervision. 
Our work is also inspired and builds upon recent works on learning geometrically interpretable optical flow fields for point cloud prediction in time \cite{SE3-Nets} and  backpropagating through  camera projection for 3D human pose estimation \cite{interpreter} or single-view depth estimation \cite{garg2016unsupervised, other_sfm}.

In summary, our contributions are: 
\begin{itemize}
\item A  method for self-supervised learning in videos \emph{in-the-wild}, through explicit modeling of the geometry of scene motion and image formation. 
\item{A deep network that predicts pixel-wise depth from a single frame along with camera motion, object motion, and object masks directly from a pair of frames.}
\item Forward-backward constraints for learning a consistent 3D structure from frame to frame and better exploit self-supervision, extending left-right consistency constraints of \cite{monodepthlr2016}.
\end{itemize}

We show results of our approach on KITTI \cite{Geiger2012CVPR, Menze2015CVPR}, MoSeg \cite{springerlink:10.1007/978-3-642-15555-0_21}, and  RGB-D SLAM \cite{sturm12iros} benchmarks under different levels of supervision.
SfM-Net learns to predict structure, object, and camera motion by training on realistic video sequences using limited ground-truth annotations.

\section{Related work}
\label{sec:related}

\paragraph{Back-propagating through warps and camera projection.}
Differentiable warping~\cite{stn} has been used to learn end-to-end unsupervised optical flow~\cite{back_to_basics:2016}, disparity flow in a stereo rig~\cite{monodepthlr2016}  and video prediction~\cite{DBLP:journals/corr/PatrauceanHC15}.  
The closest previous works to ours are SE3-Nets~\cite{SE3-Nets}, 3D image interpreter~\cite{interpreter}, and Garg et al.'s depth CNN~\cite{garg2016unsupervised}. 
SE3-Nets~\cite{SE3-Nets} use an actuation force from a robot and an input point cloud to forecast a set of 3D rigid object motions (rotation and translations) and corresponding pixel motion assignment masks under a static camera assumption. Our work uses similar representation of pixel motion masks and 3D motions to capture the dynamic objects in the scene. However, our work differs in that 
1) we predict depth and camera motion while SE3-Nets operate on given point clouds and assume no camera motion, 2) SE3-Nets are supervised with pre-recorded  3D  optical  flow, while this work admits diverse and much weaker supervision, as well as complete lack of supervision,  3)  SE3-Nets  consider  one  frame and an action as input to predict the future motion,  while our model uses pairs of frames as input to estimate the intra-frame motion, and 4) SE3-Nets are applied to toy or lab-like setups whereas we show results on real videos.

Wu et al.~\cite{interpreter} learn 3D sparse landmark positions of chairs and human body joints from a single image by computing a simplified camera model and minimizing a camera re-projection error of the landmark positions. They use synthetic data to pre-train the 2D to 3D mapping of their network. Our work considers dense structure estimation and uses videos to obtain the necessary self-supervision, instead of static images. Garg et al.~\cite{garg2016unsupervised} also predict depth from a single image, supervised by photometric error. However, they do not infer camera motion or object motion, instead requiring stereo pairs with known baseline during training.

Concurrent work to ours~\cite{other_sfm} removes the constraint that the ground-truth pose of the camera be known at training time, and instead estimates the camera motion between frames using another neural network. Our approach tackles the more challenging problem of simultaneously estimating both camera and object motion.

\paragraph{Geometry-aware motion estimation.}
Motion estimation methods that exploit rigidity of the video scene and the geometry of image formation to impose constraints on optical flow fields have a long history in computer vision~\cite{Bergen:1992:HMM:645305.648720,609381,citeulike:2638091}. Instead of non-parametric dense flow fields~\cite{Horn:1981} researchers have proposed affine or projective transformations that better exploit the low dimensionality of rigid object motion~\cite{Nir_IJCV2008}.  When depth information is available, motions are rigid rotations and translations~\cite{Hornacek_2014_CVPR}. Similarly, direct methods for visual SLAM having RGB~\cite{schoeps14ismar} or RGBD~\cite{kerl13iros} video as input,  perform dense pixel matching from frame to frame while estimating a camera trajectory and depth of the pixels in the scene
with impressive 3D point cloud reconstructions.   

These works typically make a static world assumption, which makes them susceptible to the presence of moving objects in the scene. Instead, SfM-Net explicitly accounts for moving objects using motion masks and 3D translation and rotation prediction. 

\paragraph{Learning-based motion estimation.}
Recent works~\cite{flownet,flyingthings16,Thewlis2016} propose learning frame-to-frame motion fields with deep neural networks supervised with ground-truth motion obtained from simulation or synthetic movies. This enables efficient motion estimation that learns to deal with lack of texture using training examples rather than relying only on smoothness constraints of the motion field, as previous optimization methods~\cite{Sun:CVPR:10}.  Instead of directly optimizing over unknown motion parameters, such approaches optimize neural network weights that allow motion prediction in the presence of ambiguities in the given pair of frames.

\begin{figure*}[t!]
\centering
\includegraphics[width=0.9\linewidth]{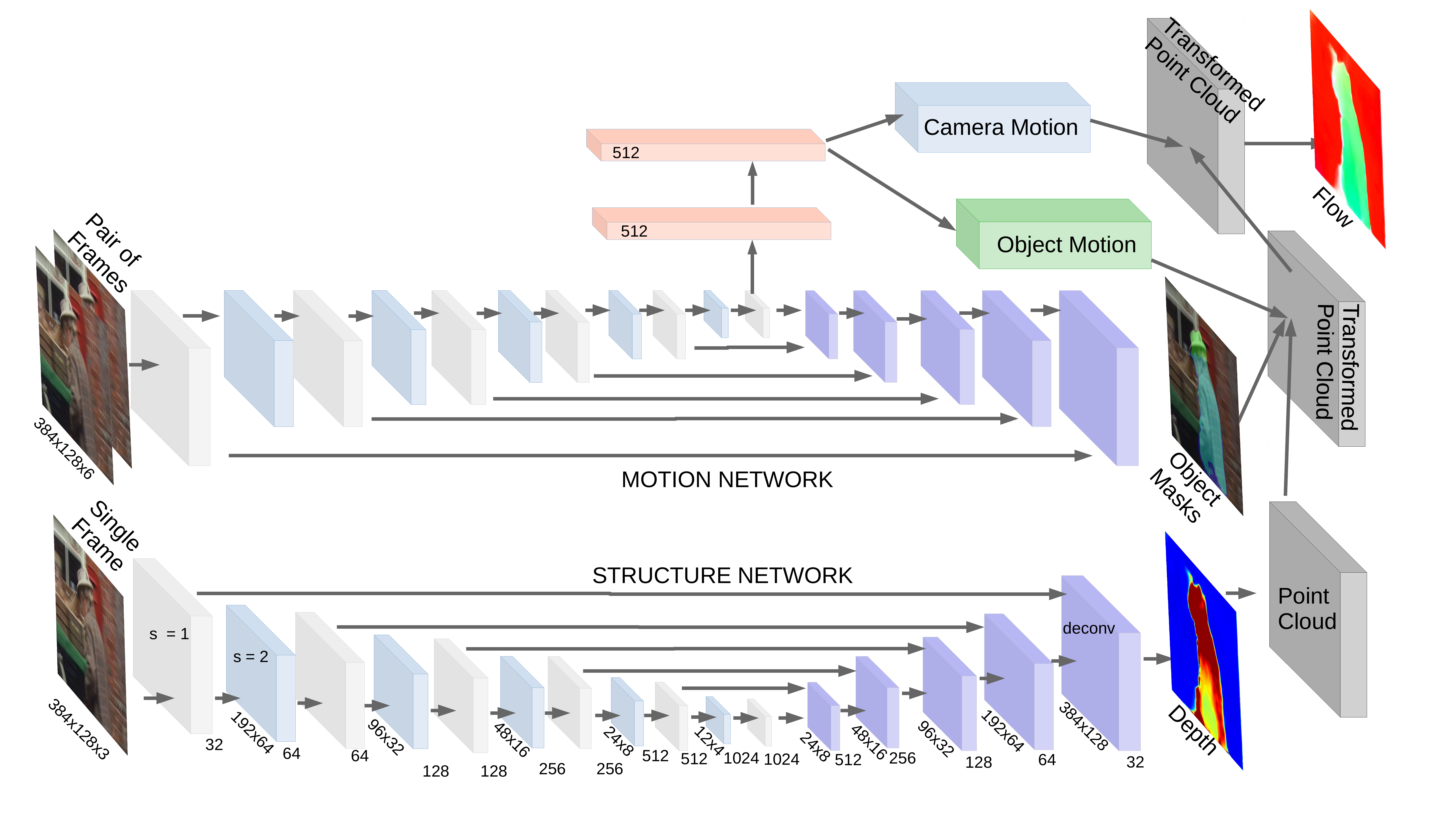}
\caption{
SfM-Net architecture. For each pair of consecutive frames $I_{t}, I_{t+1}$, a conv/deconv sub-network predicts depth $\depth_t$  while another predicts a set of $\nn$  segmentation masks $\masks_t$. The coarsest feature maps of the motion-mask encoder are further decoded through fully connected layers towards 3D rotations and translations for the camera  and the $\nn$ segmentations. The predicted depth is converted into a per frame point-cloud using estimated or known camera intrinsics. Then, it is transformed according to the predicted 3D scene flow, as composed by the 3D camera motion and independent 3D mask motions. Transformed 3D depth is projected back to the 2D next frame, and thus provides  corresponding 2D optical flow fields. Differentiable backward warping maps frame $I_{t+1}$ to $I_{t}$, and gradients are computed based on pixel errors.  Forward-backward constraints are imposed by repeating this process for the inverted frame pair $I_{t+1},I_t$ and constraining the depths $\depth_t$ and $\depth_{t+1}$ to be consistent through the estimated scene motion.
}
\label{fig:model}
\end{figure*}

\paragraph{Unsupervised learning in videos.}

Video holds a great potential towards learning semantically meaningful visual representations under weak supervision. Recent works have explored this direction by using videos to 
 propagate in time semantic labels using motion constraints \cite{conf/cvpr/PrestLCSF12}, impose temporal coherence (slowness) on the learnt visual feature  \cite{Wiskott:2002:SFA:638940.638941}, predict temporal evolution \cite{DBLP:journals/corr/WalkerDGH16}, learn temporal instance level associations \cite{DBLP:journals/corr/WangG15a}, predict temporal ordering of video frames \cite{DBLP:journals/corr/MisraZH16}, etc.  
 
Most of those unsupervised methods are shown to be good pre-training mechanisms for object detection or classification, as done in  \cite{DBLP:journals/corr/MisraZH16,DBLP:journals/corr/WalkerDGH16,DBLP:journals/corr/WangG15a}. 
In contrast and complementary to the works above, our model  extracts fine-grained 3D structure and 3D motion from monocular videos with weak supervision, instead of semantic feature representations.

\section{Learning SfM}
\label{sec:approach}
\subsection{SfM-Net architecture.}
Our model is shown in Figure \ref{fig:model}. 
Given frames $I_t$, $I_{t+1} \in  \Re^{w\times h}$, we predict frame depth $\depth_t \in [0,\infty)^{w\times h}$, camera rotation and translation $\{\cR_t, \cT_t\} \in SE3$, and a set of $\nn$ motion masks $\masks_t^k \in [0,1]^{w\times h},k \in 1, \ldots, \nn$  that denote membership of each pixel to $\nn$ corresponding rigid object motions $\{R_t^k, t_t^k\} \in SE3, k \in \{1, \ldots, \nn\}$. 
Note that a pixel may be assigned to none of the motion masks, denoting that it is a background pixel and part of the static world.
Using the above estimates, optical flow is computed by first generating the 3D point cloud corresponding to the image pixels using the depth map and camera intrinsics, transforming the point cloud based on camera and object rigid transformations, and back projecting the transformed 3D coordinates to the image plane. Then, given the optical flow field between initial and projected pixel coordinates,  differentiable backward warping is used to map frame $I_{t+1}$ to $I_{t}$.  Forward-backward constraints are imposed by repeating this process from frame $I_{t+1}$ to $I_t$ and constraining the depths $\depth_t$ and $\depth_{t+1}$ to be consistent through the estimated scene motion. 
We provide  details of each of these components below.

\paragraph{Depth and per-frame point clouds.}
We compute per frame depth using a standard conv/deconv subnetwork operating on a single frame (the structure network in Figure~\ref{fig:model}). We use a RELU activation at our final layer, since depth values are non-negative.
Given  depth $\depth_t$, we  obtain the 3D point cloud $\mathbf{X}_t^i = (X_t^i, Y_t^i, Z_t^i), i \in 1, \ldots, w \times h$  corresponding to the pixels in the scene using a pinhole camera model. Let $(x_t^i, y_t^i)$ be the column and row positions of the $i^{th}$ pixel in frame $I_t$ and let $(c_x, c_y, f)$ be  the camera intrinsics, then

\begin{footnotesize}
\begin{equation}
\mathbf{X}_t^i = 
\begin{bmatrix}
X_t^i \\
Y_t^i \\
Z_t^i \\
\end{bmatrix}
=
\frac{d_t^i}{f}\begin{bmatrix}
\frac{x_t^i}{w} - c_x \\
\frac{y_t^i}{h} - c_y\\
f \\
\end{bmatrix}
\label{eq:point_cloud}
\end{equation}
\end{footnotesize}

\noindent where $\depth_t^i$ denotes the depth value of the $i$th pixel.
We use the camera intrinsics when available and revert to default values of $(0.5, 0.5, 1.0)$ otherwise. Therefore, the predicted depth
will only be correct up to a scalar multiplier. 

\paragraph{Scene motion.}
We compute the motion of the camera and of independently moving objects in the scene using a conv/deconv subnetwork that operates on a pair of images (the motion network in Figure~\ref{fig:model}). We depth-concatenate the pair of frames and use a series of convolutional layers to produce an embedding layer. We use two fully-connected layers to predict the motion of the camera between the frames and a predefined number $\nn$ of rigid body motions that explain moving objects in the scene. 

\label{sub:camera}
Let $\{ \cR_t, \cT_t \} \in SE3$ denote the 3D rotation and translation of the camera 
from frame $I_t$ to frame $I_{t+1}$ (\textit{relative} camera pose across consecutive frames). 
We represent $\cR_{t}$ using an Euler angle representation as ${\cR_t}^{x}(\alpha){\cR_t}^{y}(\beta){\cR_t}^{z}(\gamma)$ where

{\footnotesize
\[
{\cR_t}^x(\alpha) = 
\begin{pmatrix}
\cos \alpha & -\sin \alpha & 0 \\
\sin \alpha & \cos \alpha & 0\\
0 & 0 & 1\\
\end{pmatrix},
\]
\[
{\cR_t}^y(\beta) = 
\begin{pmatrix}
\cos \beta & 0 & \sin \beta \\
0 & 1 & 0\\
-\sin \beta & 0 & \cos \beta\\
\end{pmatrix},
\]
\[
{\cR_t}^z(\gamma) = 
\begin{pmatrix}
1 & 0 & 0 \\
0 & \cos \gamma & -\sin \gamma\\
0 & \sin \gamma & \cos \gamma\\
\end{pmatrix},
\]
}

\noindent
and $\alpha, \beta, \gamma$ are the angles of rotation about the $x, y, z$-axes respectively. 
The fully-connected layers are used to predict translation parameters $\cT$, the pivot points of the camera rotation $p_c \in \mathbb{R}^3$ as in~\cite{SE3-Nets}, and   
$\sin \alpha, \sin \beta, \sin \gamma$. These last three parameters are constrained 
 
to be in the interval $[-1, 1]$ 
by using RELU activation and the minimum function.

Let $\{R_t^k, t_t^k\} \in SE3, k \in \{1, ..., \nn \}$ denote the 3D rigid motions of up to $\nn$ objects in the scene. We use similar representations as for camera motion and predict parameters using fully-connected layers on top of the same embedding $E$.  
While camera motion
is a global transformation applied to all the pixels in the scene, the object motion transforms are
weighted by the predicted membership probability of each pixel to each rigid motion, $\masks_t^k \in [0, 1]^{(h \times w)}, k \in \{1, \ldots, \nn \}$. These masks are produced by feeding the embedding layer through a deconvolutional tower. We use \emph{sigmoid} activations at the last layer instead of \emph{softmax} in order to allow each pixel to belong to any number of rigid body motions. When a pixel has zero activation across all $\nn$ maps it is assigned to the static background whose motion is a function of the global camera motion alone. We allow a pixel to belong to multiple rigid body transforms in order to capture composition of motions, e.g., through kinematic chains, such as articulated bodies. Learning the required number of motions for a sequence is an interesting open problem. We found that we could fix $\nn = 3$ for all experiments presented here. Note that our method can learn to ignore unnecessary object motions in a sequence by assigning no pixels to the corresponding mask.

\paragraph{Optical flow.}
We obtain optical flow by first transforming the point cloud obtained in Equation~\ref{eq:point_cloud} using the camera and object motion rigid body transformations followed by projecting the
3D point on to the image plane using the camera intrinsics. In the following, we drop the pixel superscript $i$ from the 3D coordinates, since it is clear we are referring to the motion transformation of the $i$th pixel of the $t$th frame. 
We first apply the object transformations:  

$\mathbf{X}'_t = \mathbf{X}_t + \sum_{k = 1}^{\nn} \mathrm{m}_t^k(i) (R_t^k (\mathbf{X}_t - p_k) + t_t^k - \mathbf{X}_t). $
We then apply the camera transformation:

$\mathbf{X}''_t = \cR_t (\mathbf{X}'_t - p^c_t) + \cT_t . $

Finally we obtain the row and column position of the pixel in the second frame $(x_{t+1}^i, y_{t+1}^i)$ by projecting the corresponding 3D point $\mathbf{X}''_t = (X''_t, Y''_t, Z''_t)$ back to the image plane as follows:
\begin{equation}
\begin{bmatrix}
\frac{x_{t+1}^i}{w} \\
\frac{y_{t+1}^i}{h} \\
\end{bmatrix}
=
\frac{f}{Z''_t}
\begin{bmatrix}
X''_t\\
Y''_t\\
f \\
\end{bmatrix}
+
\begin{bmatrix}
c_x \\
c_y \\
\end{bmatrix}
\nonumber
\end{equation}
The flow $U,V$ between the two frames at pixel $i$  is then $(U_t(i),V_t(i))=(x^i_{t+1} - x^i_t, y^i_{t+1} - y^i_t)$.

\subsection{Supervision}
SfM-Net  inverts  the image formation and extracts depth, camera and object motions that gave rise to the observed temporal differences, similar to previous SfM works \cite{SheikhNIPS2008,10.1109/ICCV.1995.466815}. Such inverse problems are ill-posed as many solutions of depth, camera and object motion can give rise to the same observed frame-to-frame pixel values.  A learning-based solution, as opposed to direct optimization, has the advantage of \textit{learning to handle such ambiguities} through partial supervision of their weights or appropriate pre-training, or simply because the same coefficients (network weights) need to explain a large abundance of video data consistently.  We detail the various supervision modes below 
and explore a subset of them in the experimental section.

\paragraph{Self-Supervision.}
Given unconstrained video, without accompanying ground-truth structure or motion information, our model is trained to  minimize the photometric error between the first frame and the second frame warped towards the first according to the predicted  motion field, based on  well-known brightness constancy assumptions~\cite{Horn:1981}: 
 
\begin{eqnarray*}
   \mathcal{L}_t^\text{color} = \frac{1}{w~h}\sum_{x,y} \|I_t(x, y) -  I_{t+1}(x', y') \|_1 \nonumber
\end{eqnarray*}
where $x' = x + U_t(x,y)$ and $y' = y + V_t(x,y)$.
We use 
 differentiable image warping proposed in the spatial transformer work~\cite{stn} and compute 
color constancy loss in a fully differentiable manner.

\paragraph{Spatial smoothness priors.}
When our network is self-supervised, we add robust spatial smoothness penalties on the optical flow field, the depth, and the inferred motion maps, by penalizing the L1 norm of the gradients  across adjacent pixels, as usually done in previous works \cite{Kong:ICCV:2015}. For  depth prediction, we penalize the norm of second order gradients in order to encourage not constant but rather smoothly changing depth values.

\paragraph{Forward-backward consistency constraints.}
We  incorporate forward-backward consistency constraints between inferred scene depth in different frames as follows. Given inferred depth $\depth_t$ from frame pair $I_t,I_{t+1}$ and  $\depth_{t+1}$ from frame pair $I_{t+1},I_{t}$, we ask for those to be consistent under the inferred scene motion, that is:
\begin{eqnarray*}
   \mathcal{L}_t^{FB} = &\frac{1}{w~h}\displaystyle \sum_{x ,y} |\left(\mathrm{d}_t(x, y)+W_t(x,y) \right) - \\ &\mathrm{d}_{t+1}(x + U_t(x, y), y + V_t(x, y))| \nonumber
\end{eqnarray*}
where $W_t(x,y)$ is the $Z$ component of the scene flow obtained from the point cloud transformation.
Composing scene flow forward and backward across consecutive frames allows us to impose such forward-backward consistency cycles across more than one frame gaps, however, we have not yet seen empirical gain from doing so.

\paragraph{Supervising depth.}
If depth is available on parts of the input image, such as with video sequences captured by a Kinect sensor,  we can use depth supervision in the form of robust depth regression:
\begin{equation}
   \mathcal{L}_t^{depth} = \frac{1}{w~h}\sum_{x,y} \mathrm{dmask}^{GT}_t(x,y) \cdot \| \mathrm{d}_t(x,y) - \mathrm{d}^{GT}_t(x,y) \| _1, \nonumber
\end{equation}
where $\mathrm{dmask}^{GT}_t$ denotes a binary image that signals presence of ground-truth depth.

\paragraph{Supervising camera motion.}
If ground-truth camera pose trajectories are available, we can supervise our model by computing corresponding ground-truth camera rotation and translation $R_t^{c-GT},t_t^{c-GT}$ from frame to frame, and constrain our camera motion predictions accordingly.  Specifically, we compute the relative transformation between predicted and ground-truth camera motion $\{t^\text{err}_t=\mathrm{inv}(R^c_t)(\mathrm{t}^{c-GT}_t - \mathrm{t}^{c}_t), R^\text{err}_t=\mathrm{inv}(R_t^{c})R_t^{c-GT})  \}$
and minimize its rotation angle and translation norm \cite{sturm12iros}:
{\small
\begin{equation}
\begin{array}{r@{}l}
   \mathcal{L}_t^{c_\text{trans}} &{}= \| t^\text{err}_t \|_2 \\
   \mathcal{L}_t^{c_\text{rot}} &{}= \mathrm{arccos}\left(\mathrm{min}\left(1,\mathrm{max}\left(-1,\frac{\mathrm{trace}(R^{err}_t)-1}{2}\right)\right)\right) \label{eq:cameraposerr}
\end{array}
\end{equation}
}

\paragraph{Supervising optical flow and object motion.}
 Ground-truth optical flow, object masks, or object motions require expensive human annotation on real videos. However, these signals are available in recent synthetic datasets~\cite{flyingthings16}. In such cases, our model could be trained to minimize, for example, an L1 regression loss between predicted $\{ U(x, y),V(x, y) \} $ and ground-truth $\{ U^{GT}(x, y),V^{GT}(x, y) \} $ flow vectors.

\subsection{Implementation details}
Our depth-predicting structure and object-mask-predicting motion conv/deconv networks share similar architectures but use independent weights. Each consist of a series of $3\times 3$ convolutional layers alternating between stride 1 and stride 2 followed by deconvolutional operations consisting of a depth-to-space upsampling, concatentation with corresponding feature maps from the convolutional portion, and a $3\times3$ convolutional layer. Batch normalization is applied to all convolutional layer outputs. The structure network takes a single frame as input, while the motion network takes a pair of frames. We predict depth values using a $1\times1$ convolutional layer on top of the image-sized feature map. We use RELU activations because depths are positive and a bias of $1$ to prevent small depth values. The maximum predicted depth value is further clipped at $100$ to prevent large gradients. We predict object masks from the image-sized feature map of the motion network using a $1\times1$ convolutional layer with sigmoid activations. To encourage sharp masks we multiply the logits of the masks by a parameter that is a function of the number of step for which the network has been trained. The pivot variables are predicted as heat maps using a softmax function over all the locations in the image followed by a weighted average of the pixel locations.

\section{Experimental results}
\label{sec:experiments}

The main contribution of SfM-Net is the ability to explicitly model both camera and object motion in a sequence, allowing us to train on unrestricted videos containing moving objects. To demonstrate this, we trained self-supervised networks (using zero ground-truth supervision) on the KITTI datasets~\cite{Geiger2012CVPR, Menze2015CVPR} and on the MoSeg dataset~\cite{springerlink:10.1007/978-3-642-15555-0_21}. KITTI contains pairs of frames captured from a moving vehicle in which other independently moving vehicles are visible. MoSeg contains sequences with challenging object motion, including articulated motions from moving people and animals.

\paragraph{KITTI.} 
 Our first experiment validates that explicitly modeling object motion is necessary to effectively learn from unconstrained videos. We evaluate unsupervised depth prediction using our models on the KITTI 2012 and KITTI 2015 datasets which contain close to 200 frame sequence and stereo pairs. We use a scale-invariant error metric (log RMSE) proposed in \cite{eigen} due to the global scale ambiguitiy in monocular setups which is defined as
 \[
     \mathcal{E}_{\text{scaleinv}} = \frac{1}{N} \sum_{x,y} \| \bar{d}(x,y) \| _2 - 
   \left(
    \frac{1}{N} \sum_{x,y} \| \bar{d}(x,y) \|_1 \right)^2,
\]
where $N$ is the number of pixels and $\bar{d}=(\log(\mathrm{d}) - \log(\mathrm{d}^{GT}))$ denotes the difference between the log of ground-truth and predicted depth maps. We pre-train the our unsupervised depth prediction models using adjacent frame pairs on the raw KITTI dataset which contains $\sim 42,000$ frames and train and evaluate on KITTI 2012 and 2015 which have depth ground truth. 

 We compare the the results of Garg et al.~\cite{garg2016unsupervised} who use stereo pairs to estimate depth. Their approach assumes the camera pose between the frames is a known constant (stereo baseline) and optimize the photometric error in order to estimate the depth. In contrast, our model considers a more challenging ``in the wild" setting where we are only given sequences of frames from a video and camera pose, depth and object motion are all estimated without any form of supervision. Garg et al. report a log RMSE of 0.273 on a subset of the KITTI dataset. To compare with our approach on the full set we emulate the model of Garg et al. using our architecture by removing object masks from our network and using stereo pairs with photometric error. We also evaluate our full model on frame sequence pairs with camera motion estimation both with and without explicit object motion estimation.

Table~\ref{tab:kitti_depth} shows the log RMSE error between the ground-truth depth and the three approaches. When using stereo pairs we obtain a value of $0.31$ which is on par with existing results on the KITTI benchmark (see \cite{garg2016unsupervised}). When using frame sequence pairs instead of calibrated stereo pairs the problem becomes more difficult, as we must now infer the unknown camera and object motion between the two frames. As expected, the depth estimates learned in this scenario are less accurate, but performance is much worse when no motion masks are used. The gap between the two approaches is wider on the KITTI 2015 dataset which contains more moving objects. This shows that it is important to account for moving objects when training on videos in the wild.

 \begin{table}
    \centering
    \begin{tabular}{|c|c|c|}
    \hline
    \textbf{Approach} & \multicolumn{2}{c|}{\textbf{Log RMSE}} \\
    & \textbf{KITTI 2012} & \textbf{KITTI 2015} \\
    \hline
    with stereo pairs & 0.31 & 0.34 \\
    seq.~with motion masks & 0.45 & 0.41 \\
    seq.~without motion masks & 0.77 & 1.25 \\
    \hline
    \end{tabular}
    \caption{RMSE of Log depth with respect to ground truth for our model with stereo pairs and with and without motion masks on sequences in KITTI 2012 and 2015 datasets. When using stereo pairs the camera pose between the frames is fixed and the model is equivalent to the approach of Garg et al. \cite{garg2016unsupervised}. Motion masks help improve the error on both datasets but more so on the KITTI 2015 dataset which contains more moving objects.}
    \label{tab:kitti_depth}
\end{table}

\begin{figure}
\centering
\begin{tabular}{ccc}
\textbf{RGB frame} & \multicolumn{2}{c}{\textbf{Predicted Depth}} \\
& \textbf{(stereo pairs)} & \textbf{(sequence)} \\
\includegraphics[width=0.27\linewidth]{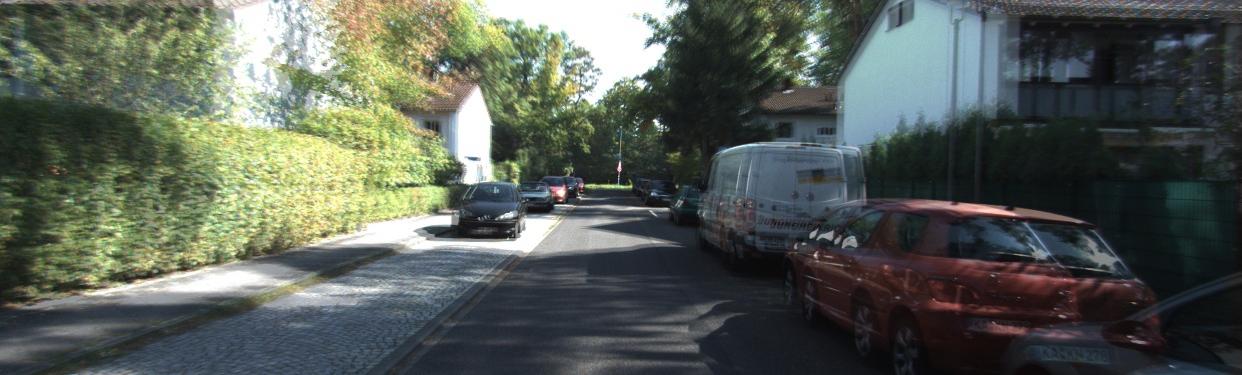} &
\includegraphics[width=0.27\linewidth]{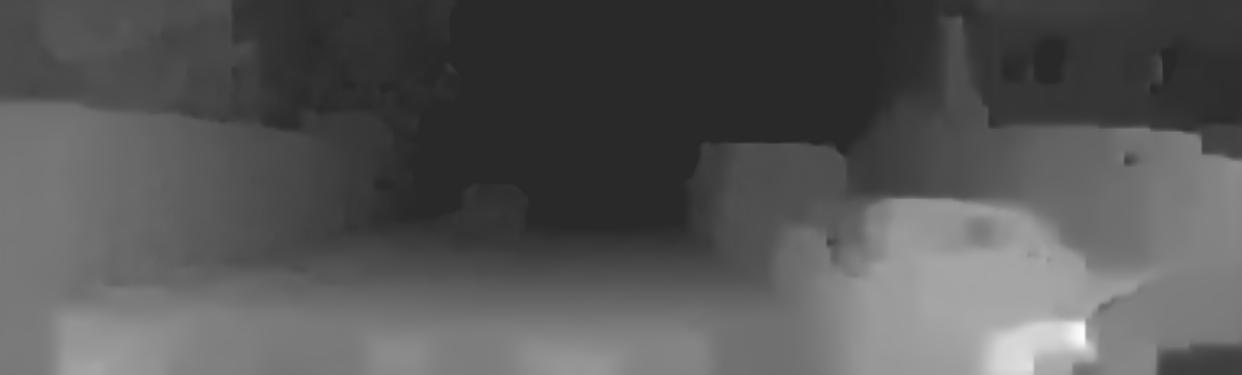} &
\includegraphics[width=0.27\linewidth]{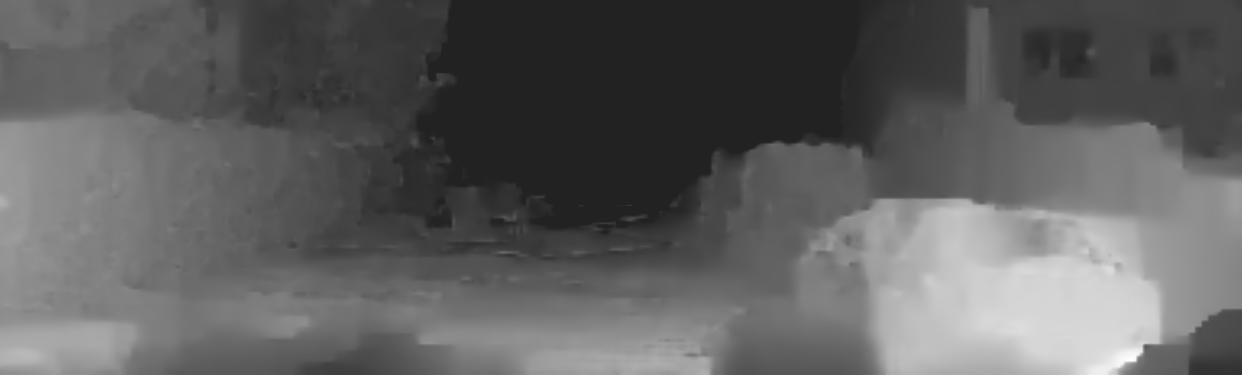} \\
\includegraphics[width=0.27\linewidth]{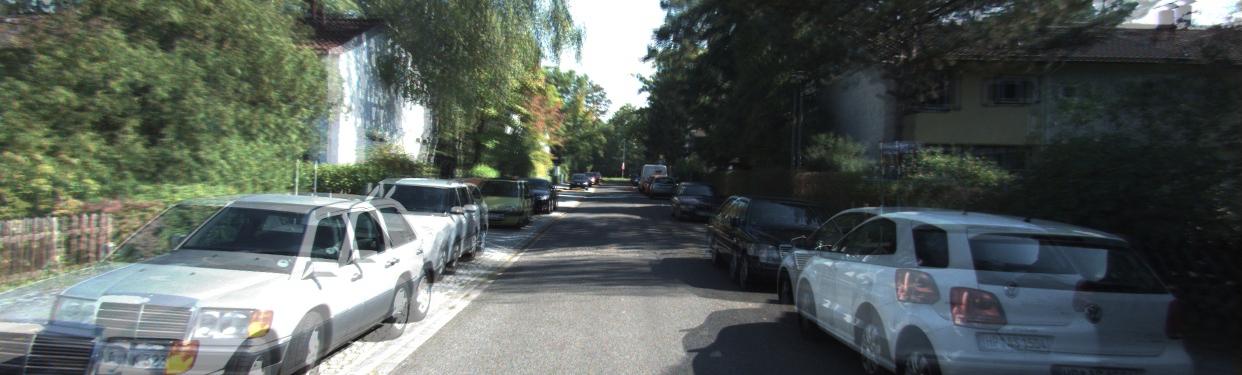} &
\includegraphics[width=0.27\linewidth]{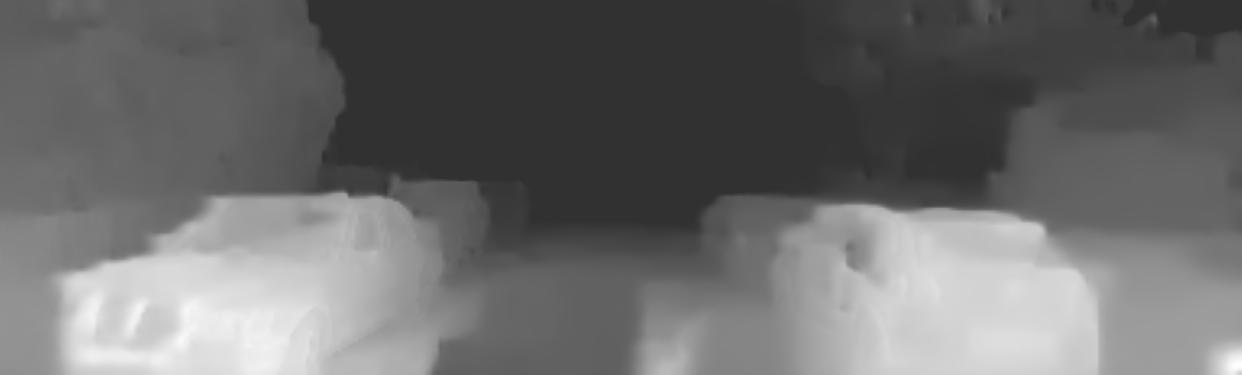} &
\includegraphics[width=0.27\linewidth]{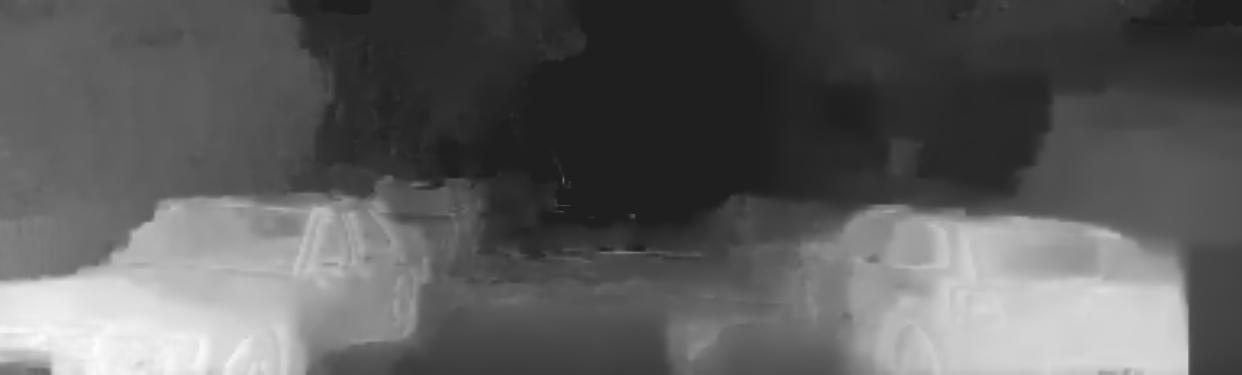} \\
\includegraphics[width=0.27\linewidth]{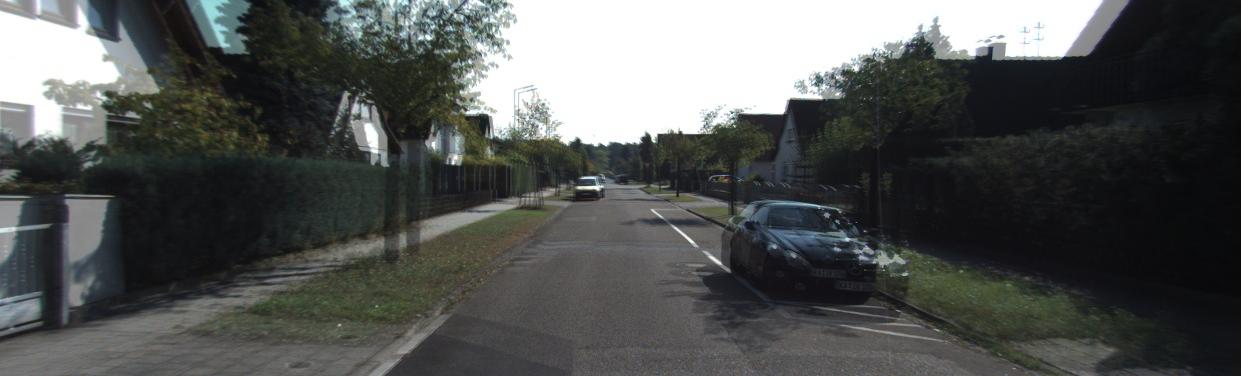} &
\includegraphics[width=0.27\linewidth]{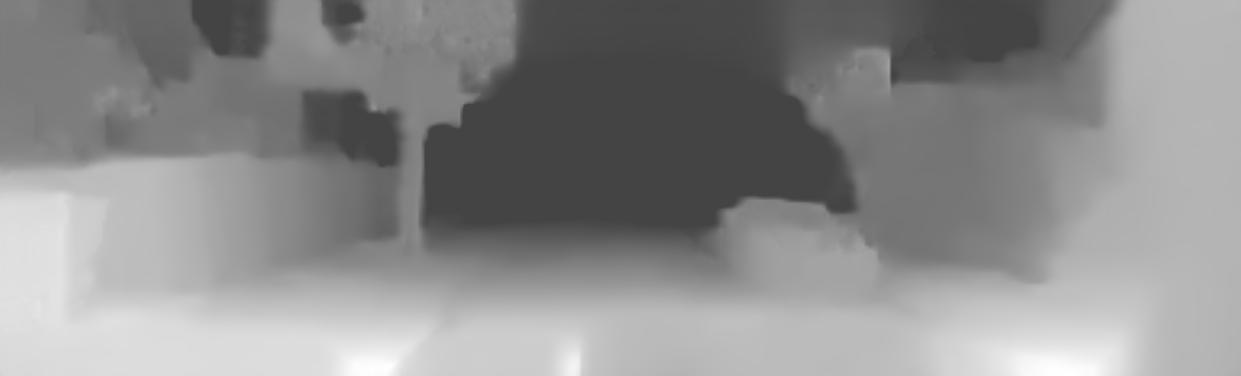} &
\includegraphics[width=0.27\linewidth]{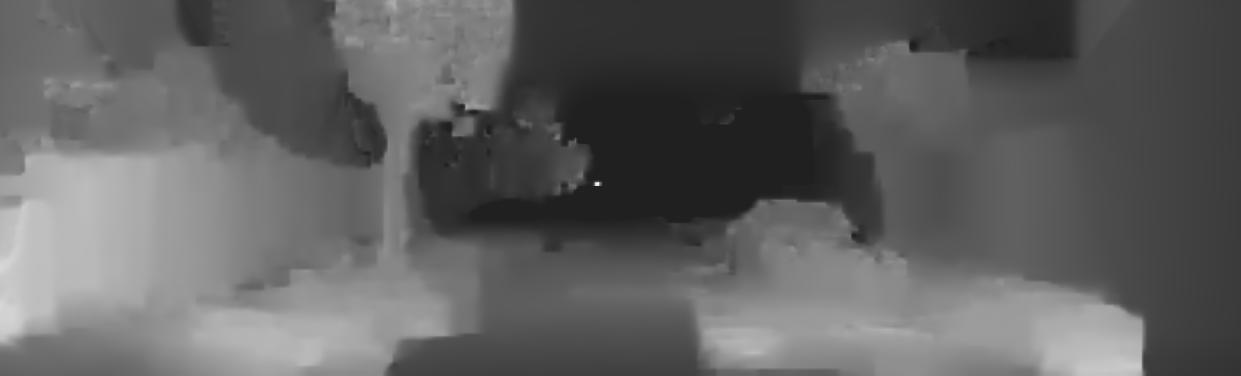} \\
\includegraphics[width=0.27\linewidth]{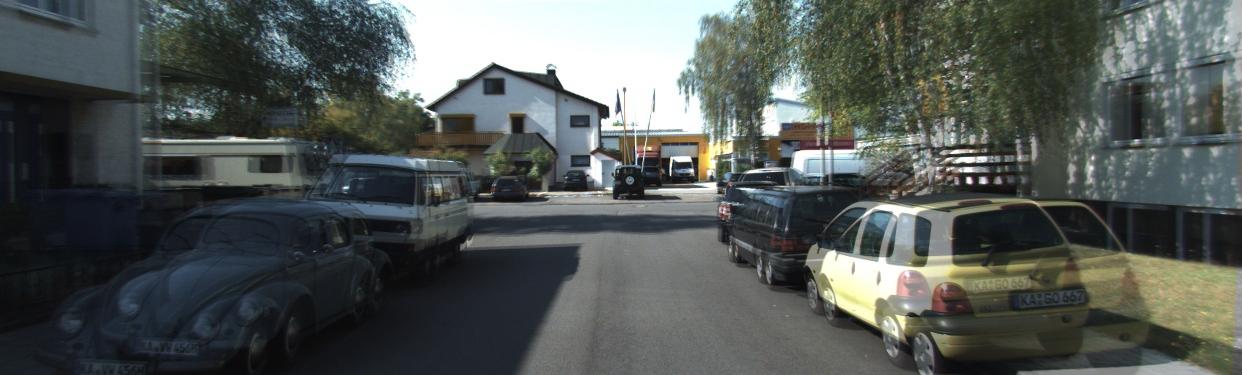} &
\includegraphics[width=0.27\linewidth]{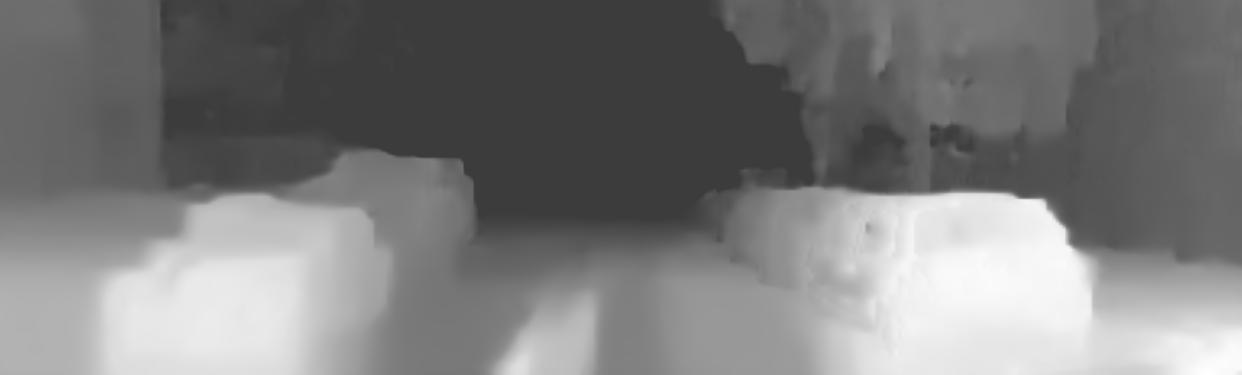} &
\includegraphics[width=0.27\linewidth]{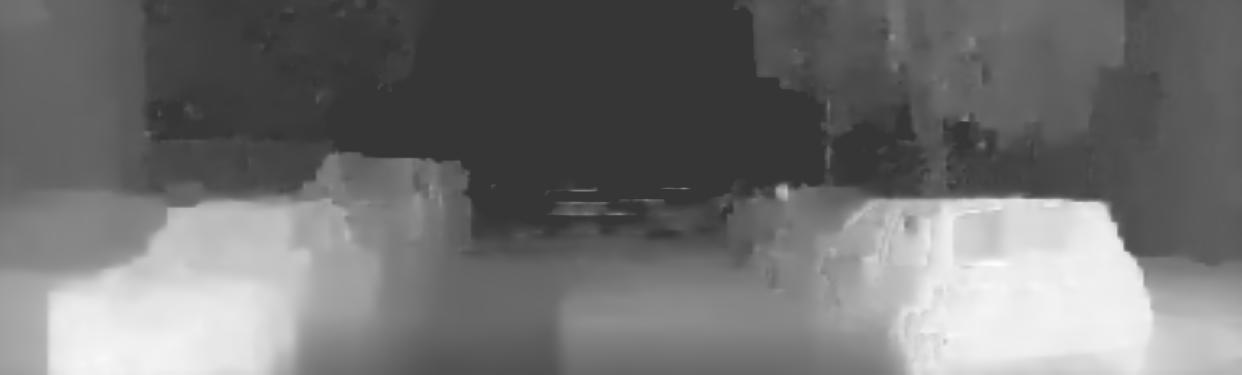} \\
\includegraphics[width=0.27\linewidth]{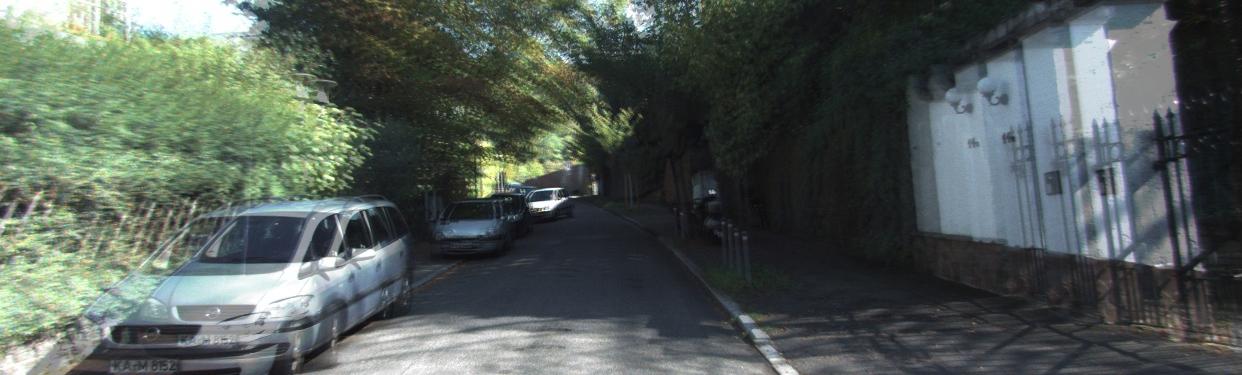} &
\includegraphics[width=0.27\linewidth]{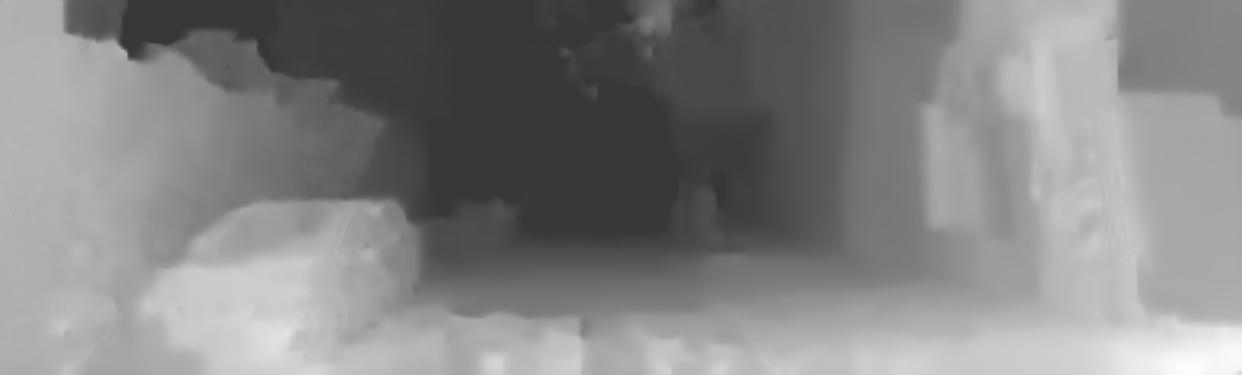} &
\includegraphics[width=0.27\linewidth]{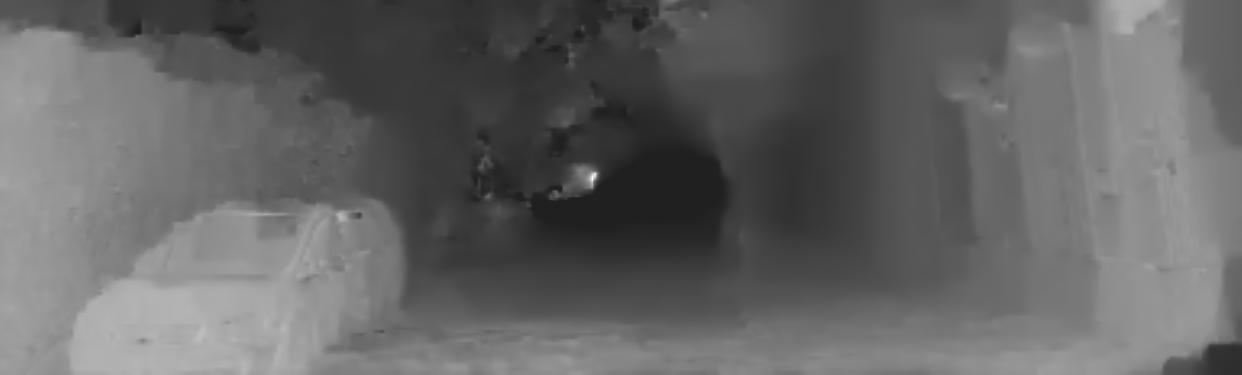} \\
\includegraphics[width=0.27\linewidth]{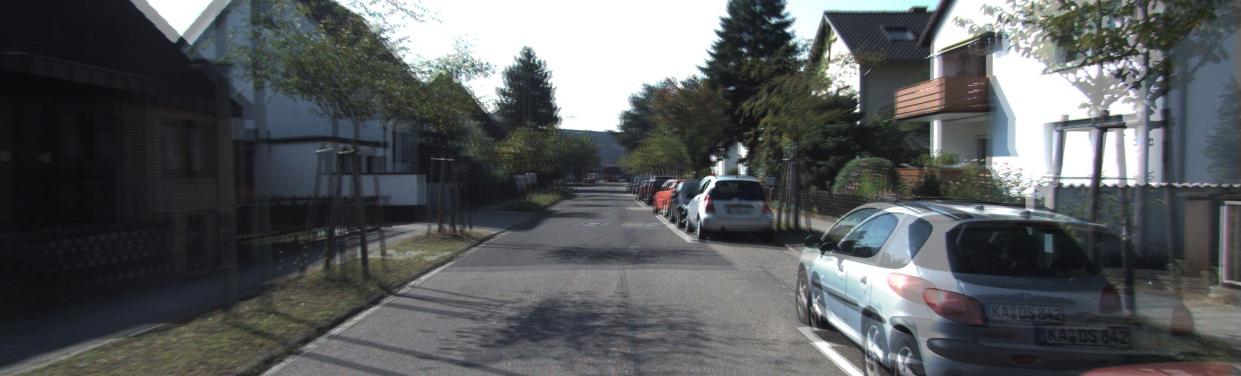} &
\includegraphics[width=0.27\linewidth]{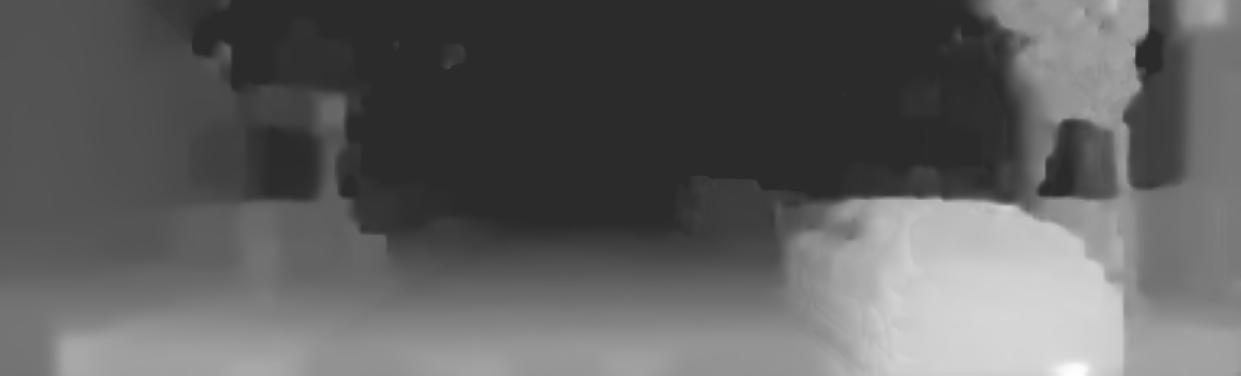} &
\includegraphics[width=0.27\linewidth]{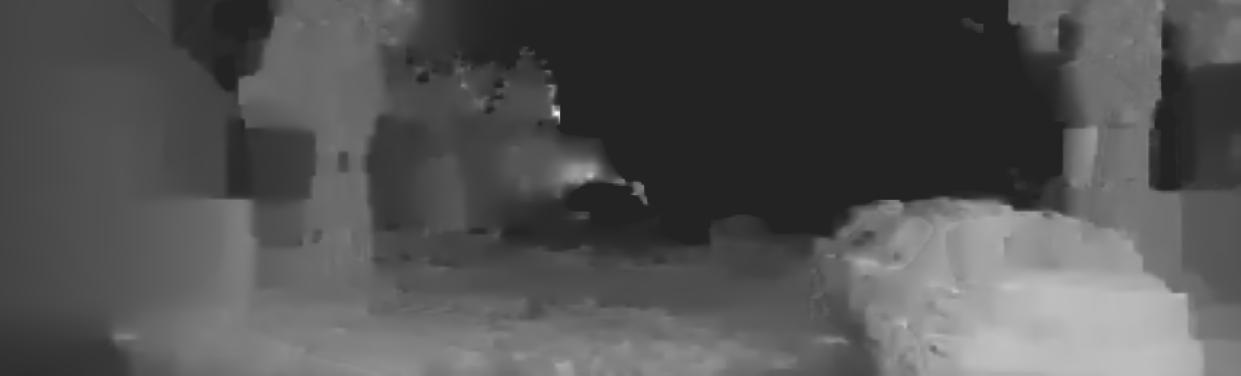} \\
\hline
\hline
\includegraphics[width=0.27\linewidth]{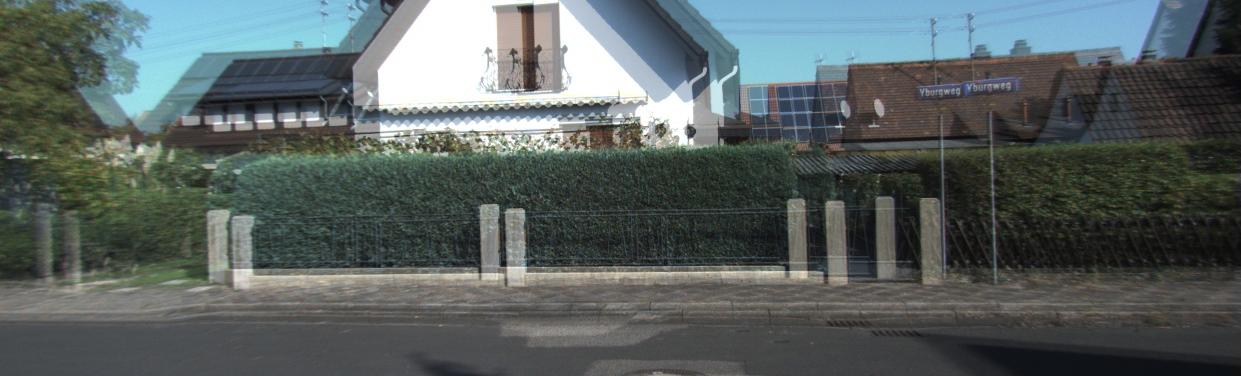} &
\includegraphics[width=0.27\linewidth]{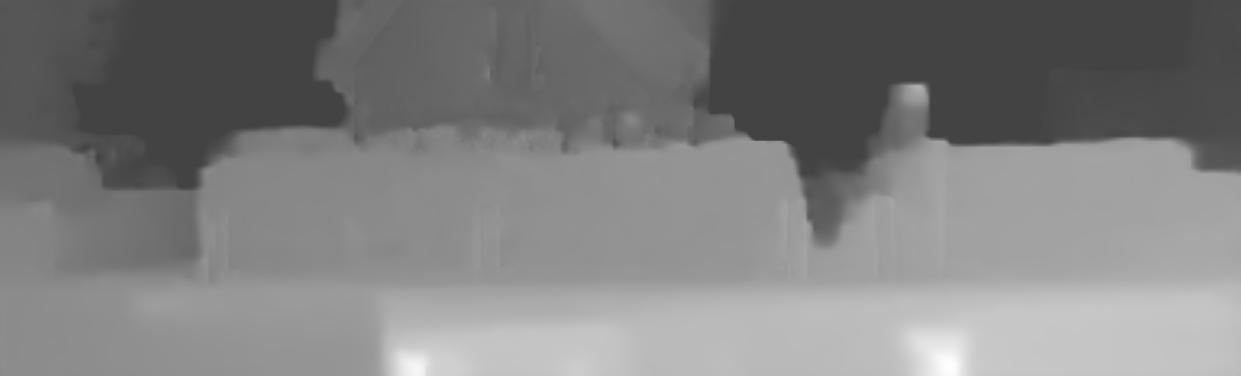} &
\includegraphics[width=0.27\linewidth]{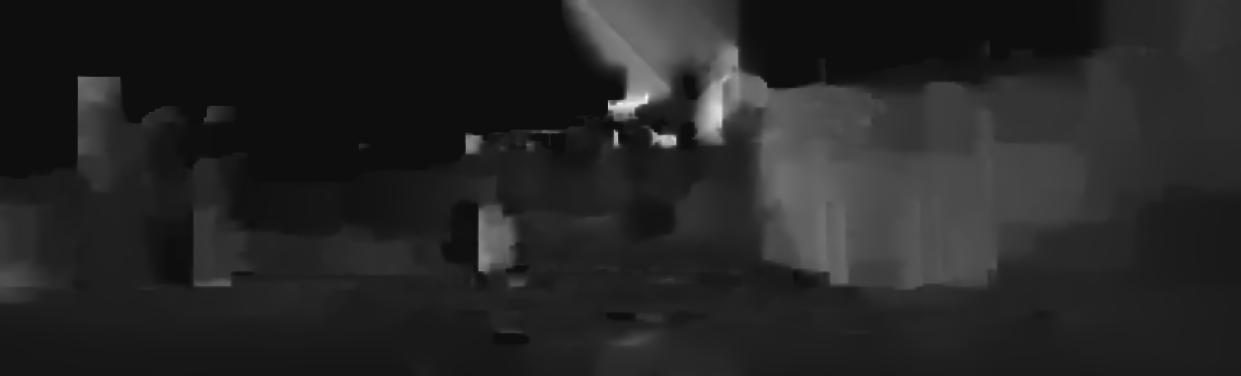} \\
\includegraphics[width=0.27\linewidth]{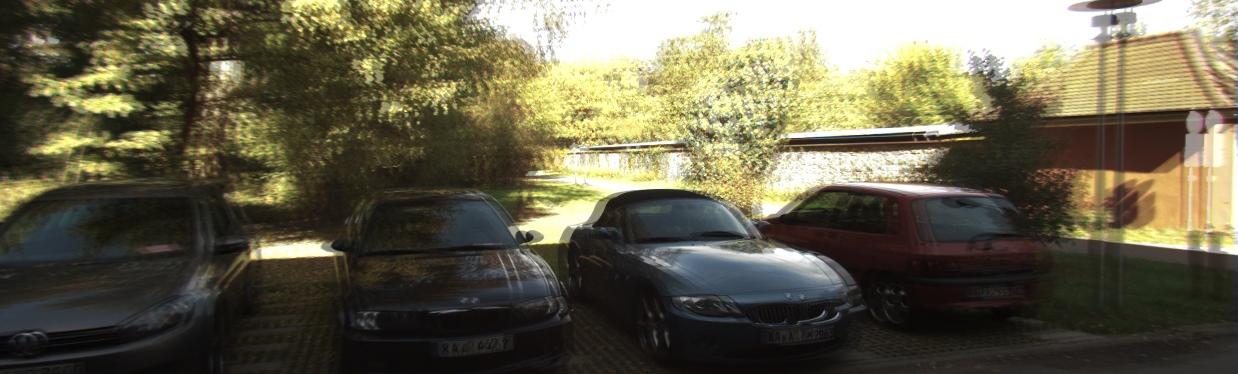} &
\includegraphics[width=0.27\linewidth]{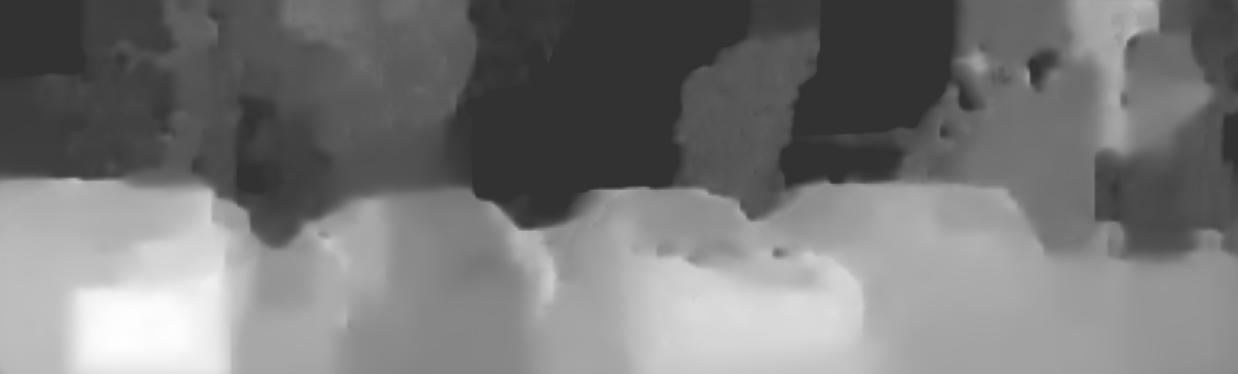} &
\includegraphics[width=0.27\linewidth]{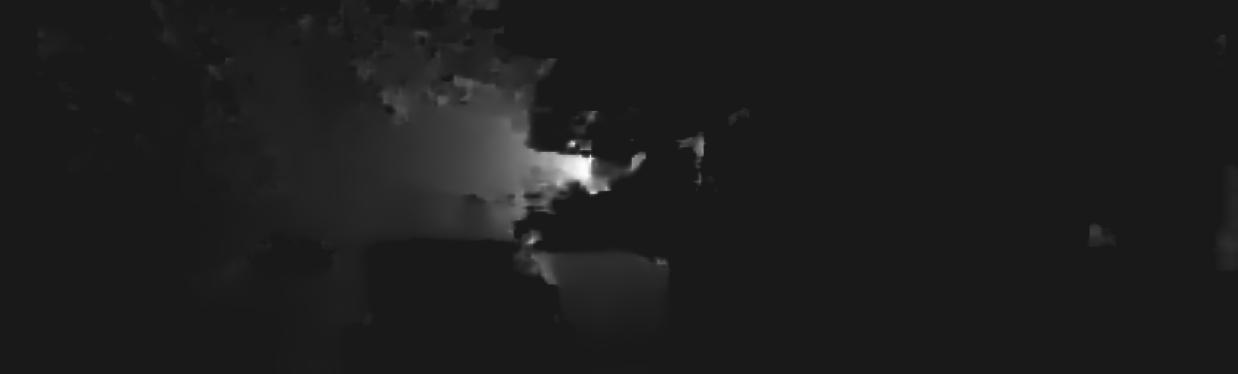} \\

\end{tabular}
\caption{Qualitative comparison of the estimated depth using our unsupervised model on sequences versus using stereo pairs in the KITTI 2012 benchmark. When using stereo pairs the camera pose between the pair is constant and hence the model is equivalent to the approach of Garg et al. \cite{garg2016unsupervised}. For sequences, our model needs to additionally predict camera rotation and translation between the two frames. The first six rows show successful predictions even without camera pose information and the last two illustrate failure cases. The failure cases show that when there is no translation between the two frames depth estimation fails whereas when using stereo pairs there is always a constant offset between the frames.}
\label{fig:kitti_depth}
\end{figure}

\begin{figure*}
\centering
\begin{tabular}{cccc}
  \textbf{Predicted Motion Masks} & \textbf{Ground Truth Mask} & \textbf{Predicted Flow} & \textbf{Ground Truth Flow} \\
\includegraphics[width=0.23\linewidth]{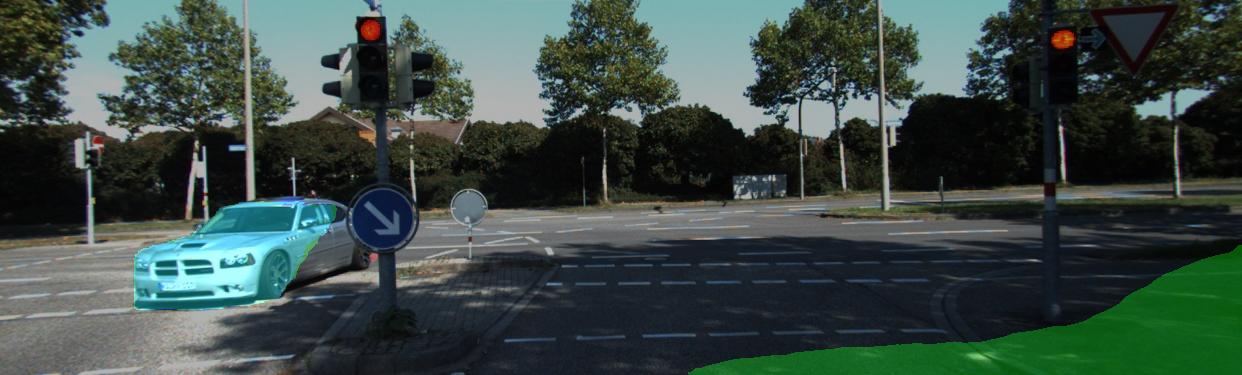} &
\includegraphics[width=0.23\linewidth]{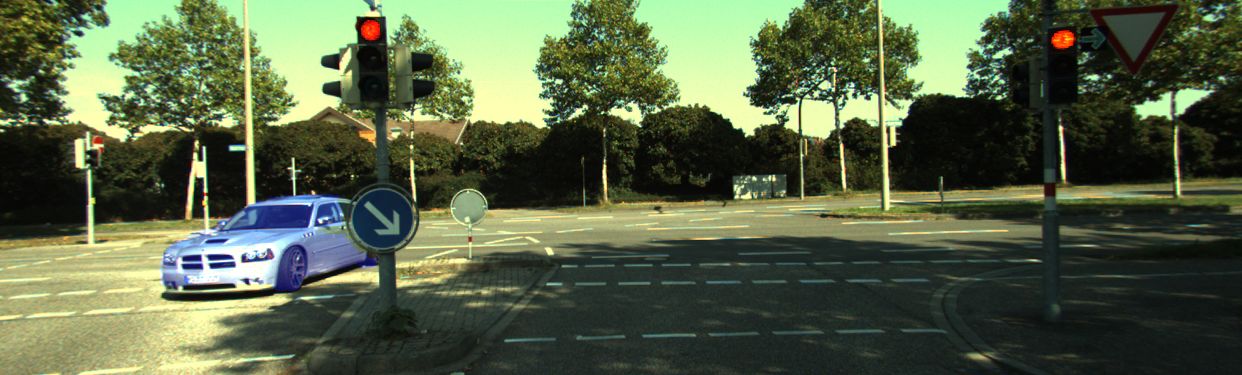} &
\includegraphics[width=0.23\linewidth]{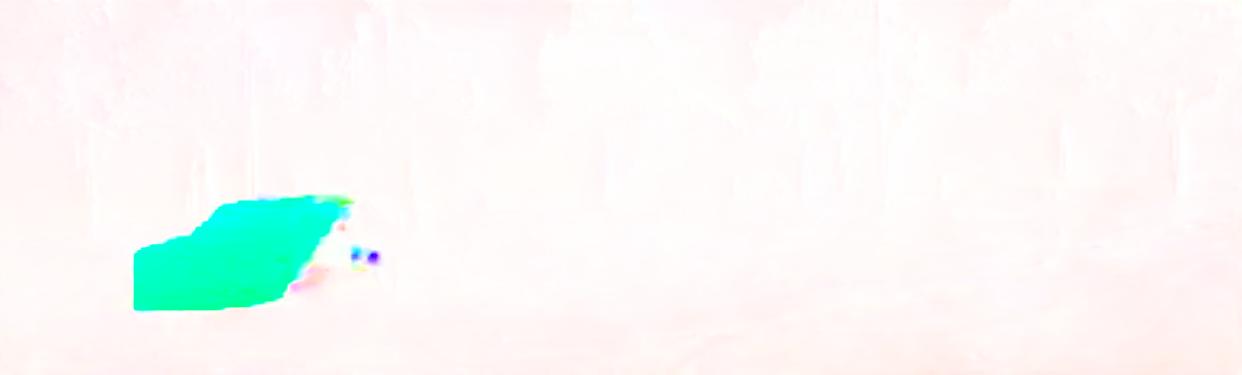} &
\includegraphics[width=0.23\linewidth]{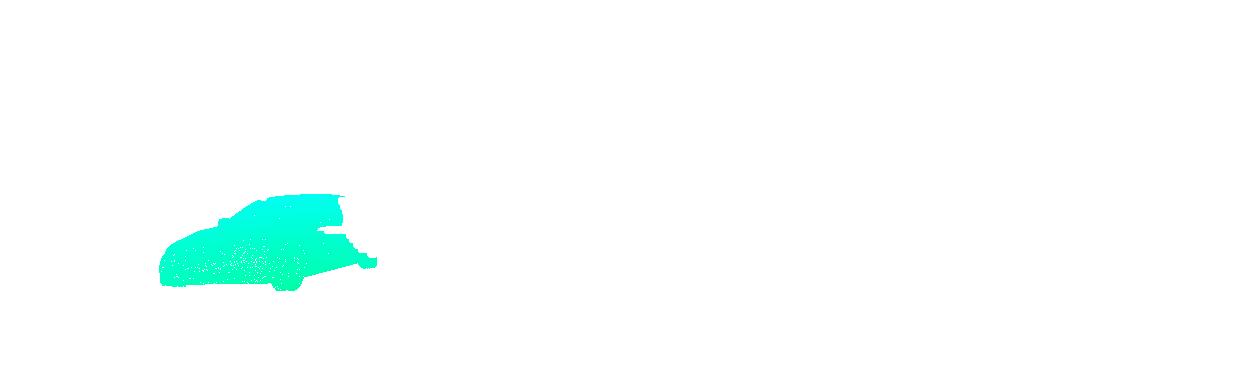} \\
\includegraphics[width=0.23\linewidth]{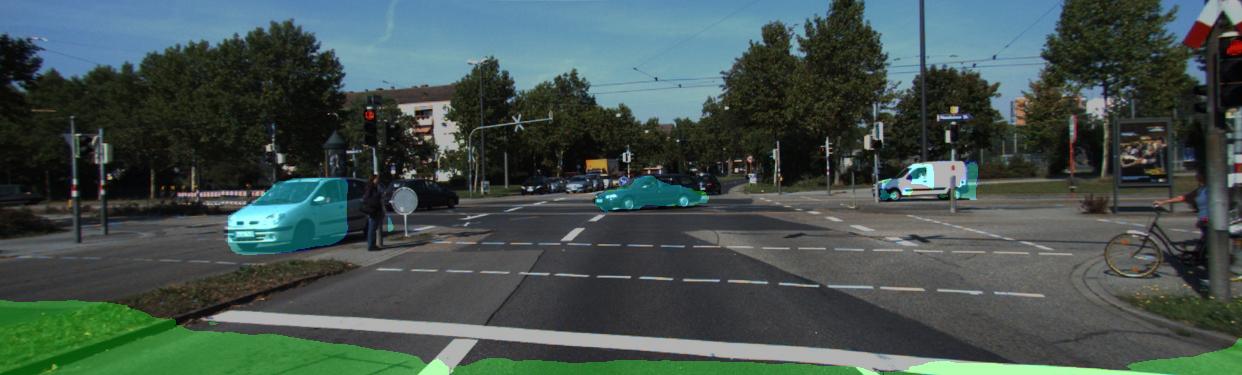} &
\includegraphics[width=0.23\linewidth]{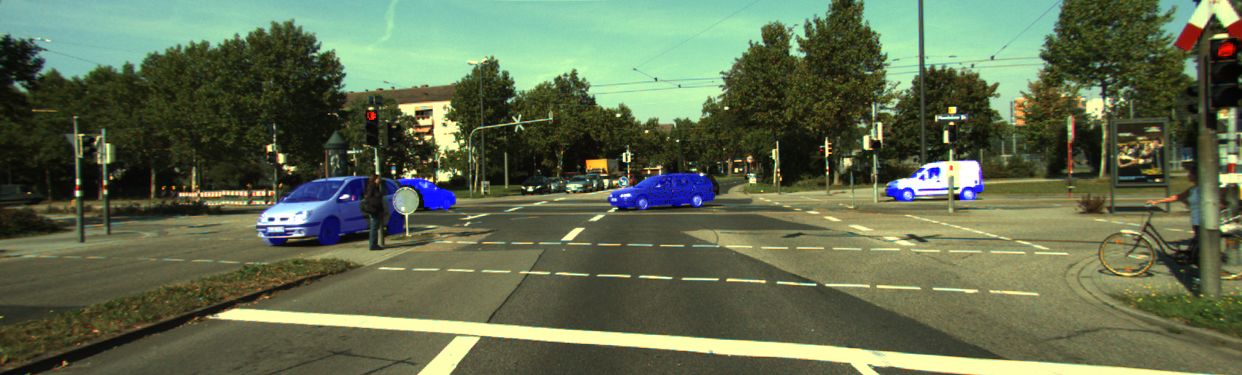} &
\includegraphics[width=0.23\linewidth]{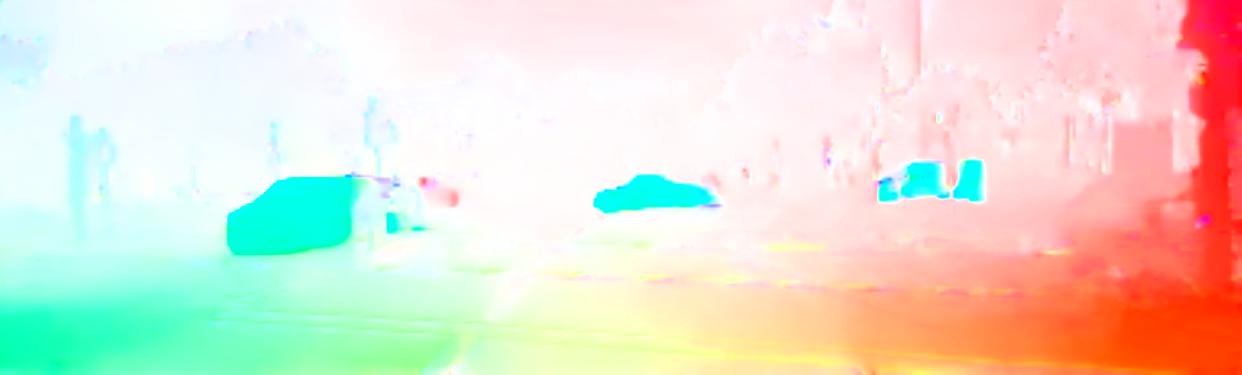} &
\includegraphics[width=0.23\linewidth]{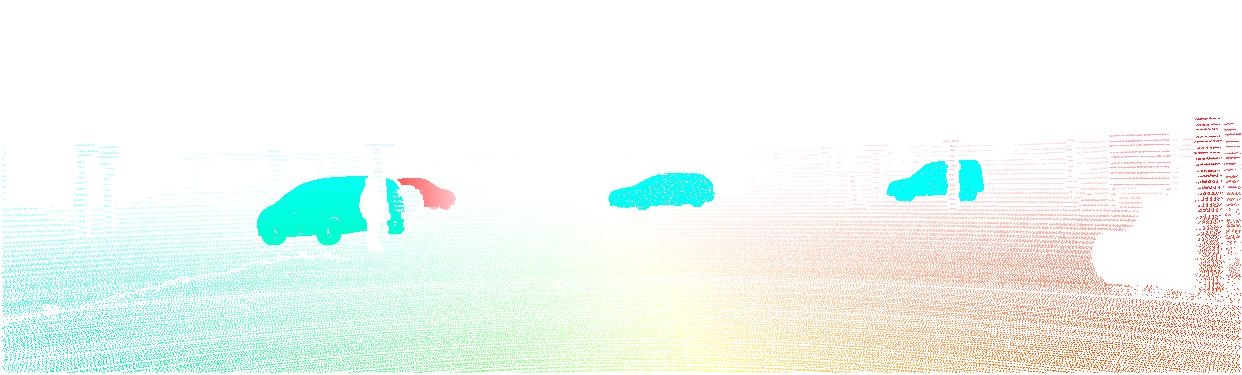} \\
\includegraphics[width=0.23\linewidth]{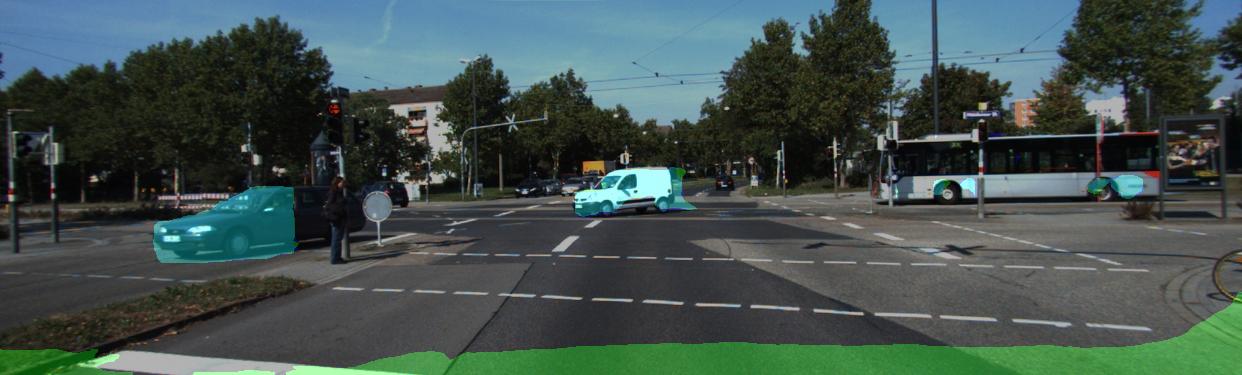} &
\includegraphics[width=0.23\linewidth]{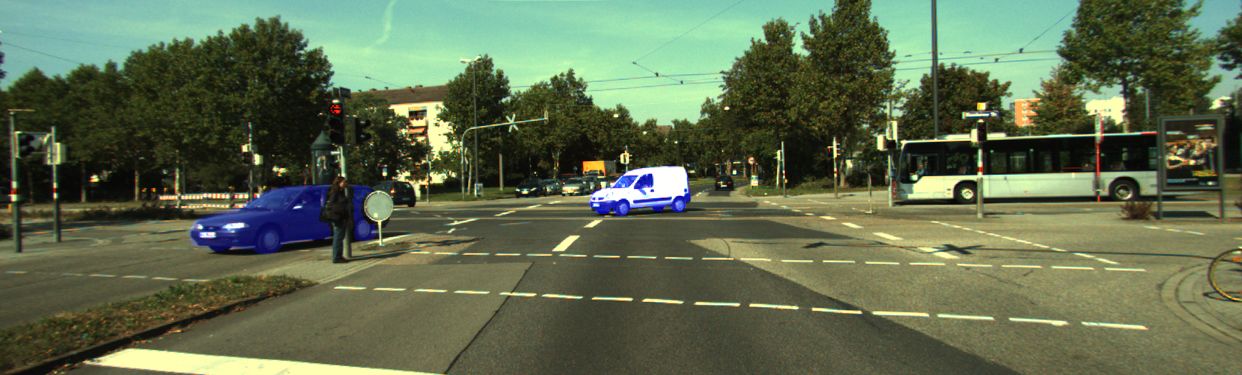} &
\includegraphics[width=0.23\linewidth]{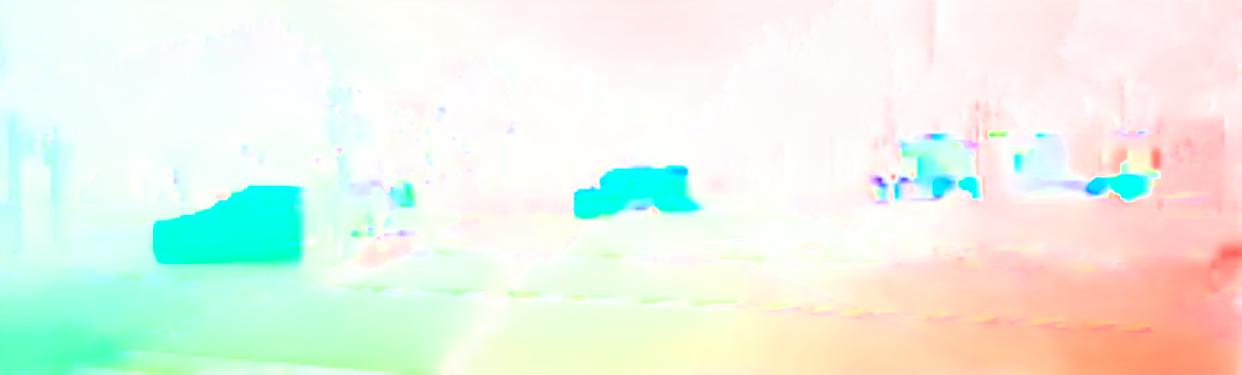} &
\includegraphics[width=0.23\linewidth]{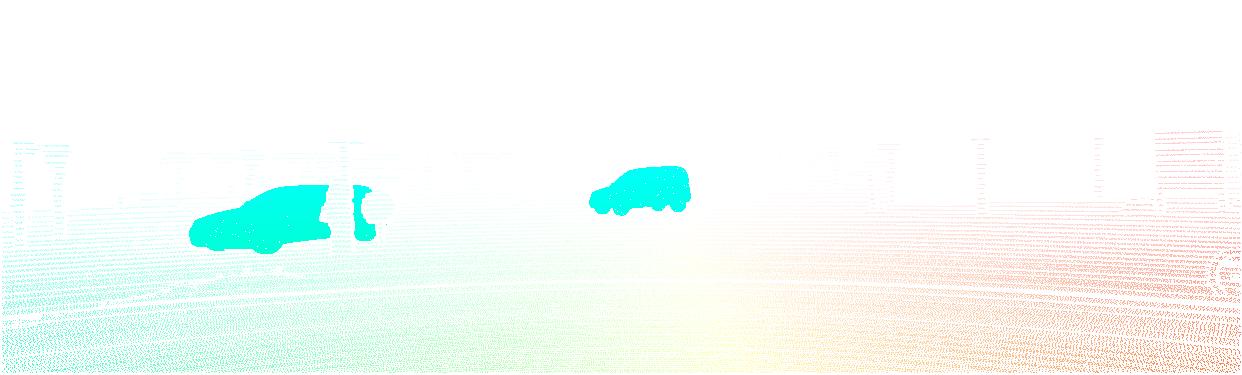} \\
\includegraphics[width=0.23\linewidth]{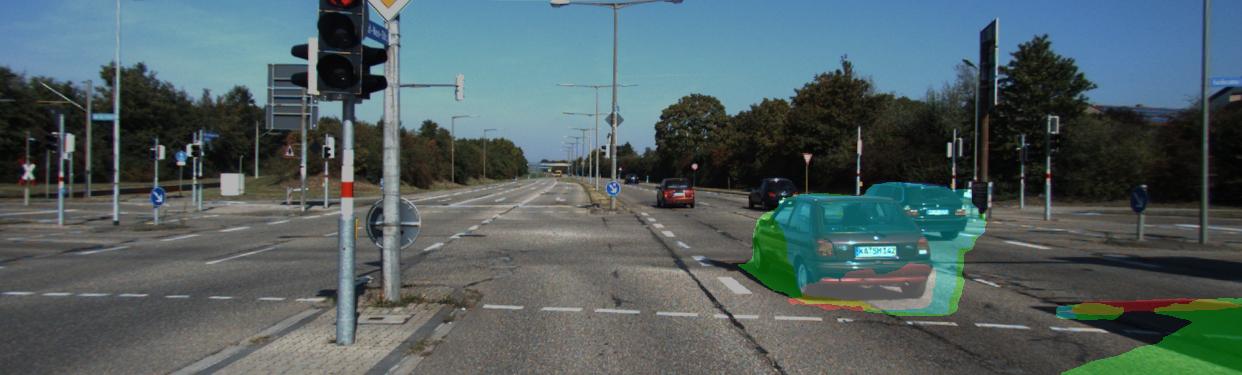} &
\includegraphics[width=0.23\linewidth]{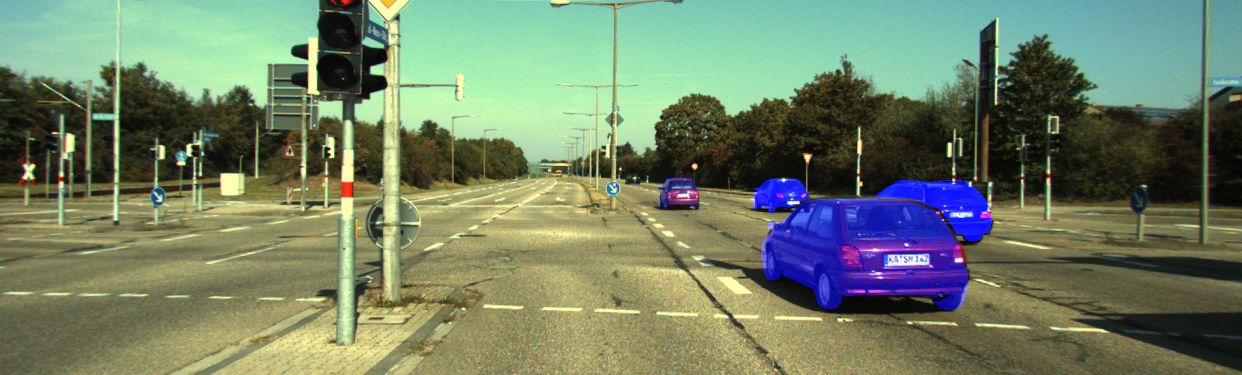} &
\includegraphics[width=0.23\linewidth]{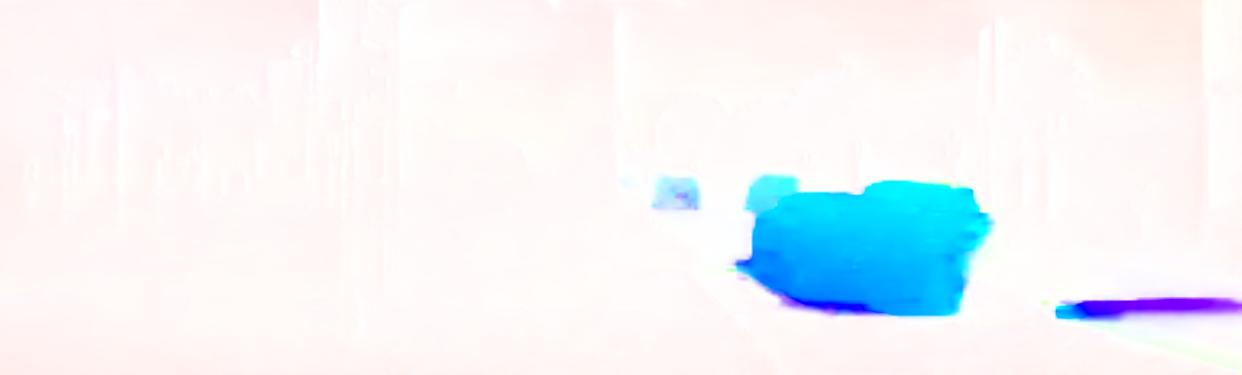} &
\includegraphics[width=0.23\linewidth]{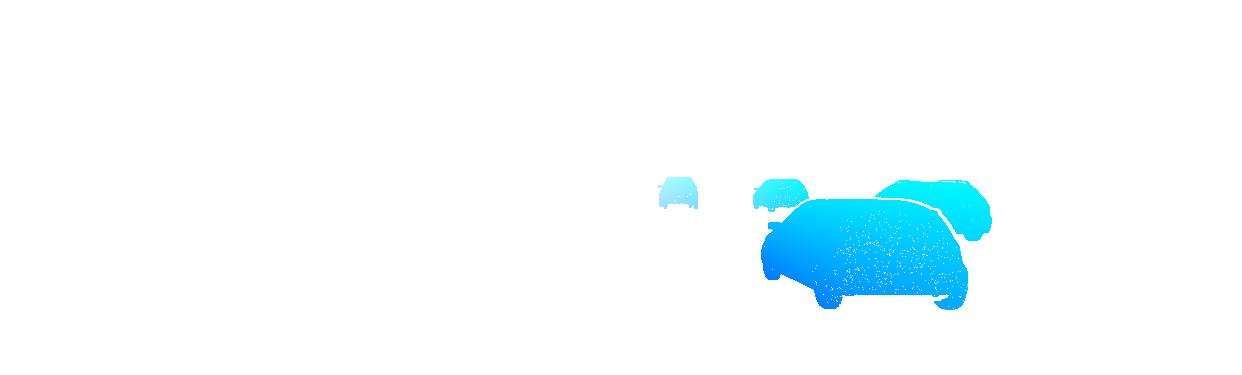} \\
\includegraphics[width=0.23\linewidth]{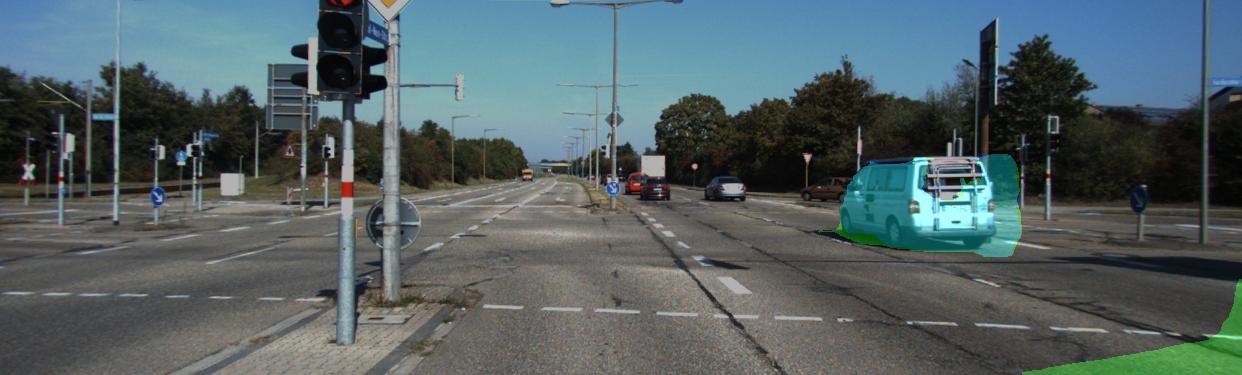} &
\includegraphics[width=0.23\linewidth]{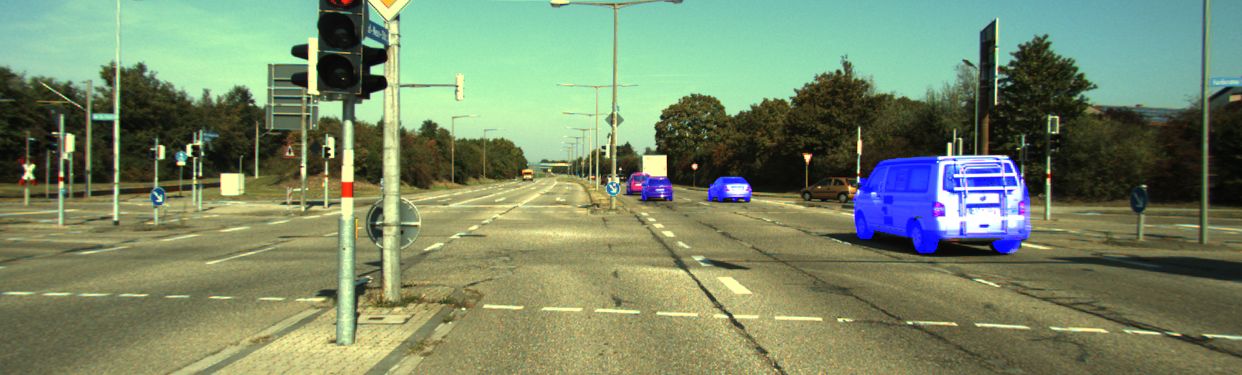} &
\includegraphics[width=0.23\linewidth]{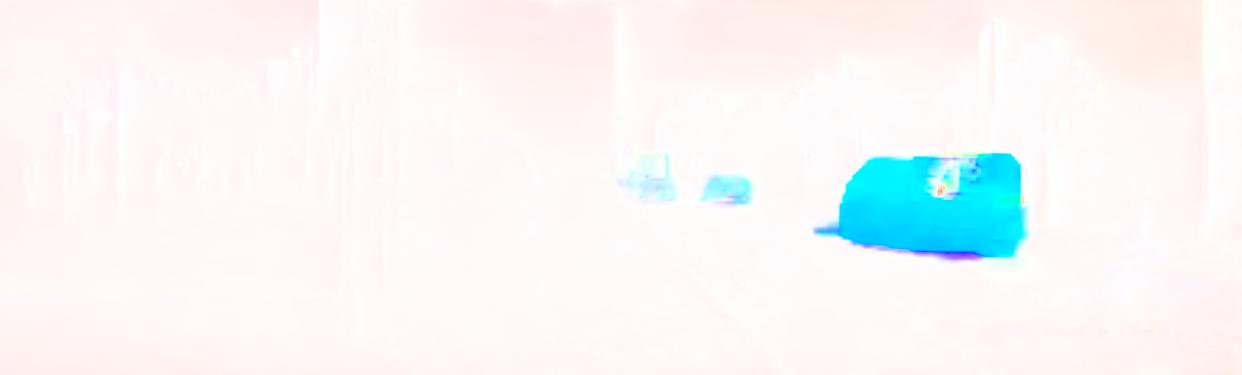} &
\includegraphics[width=0.23\linewidth]{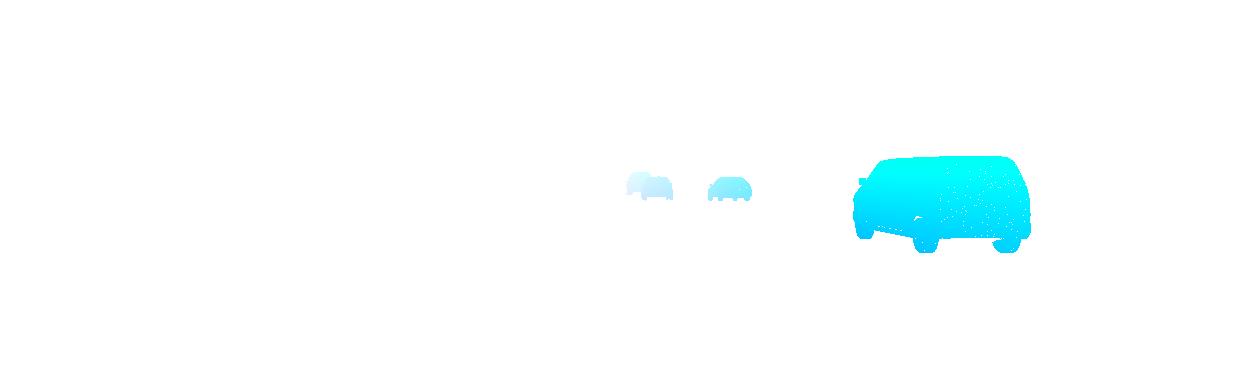} \\
\includegraphics[width=0.23\linewidth]{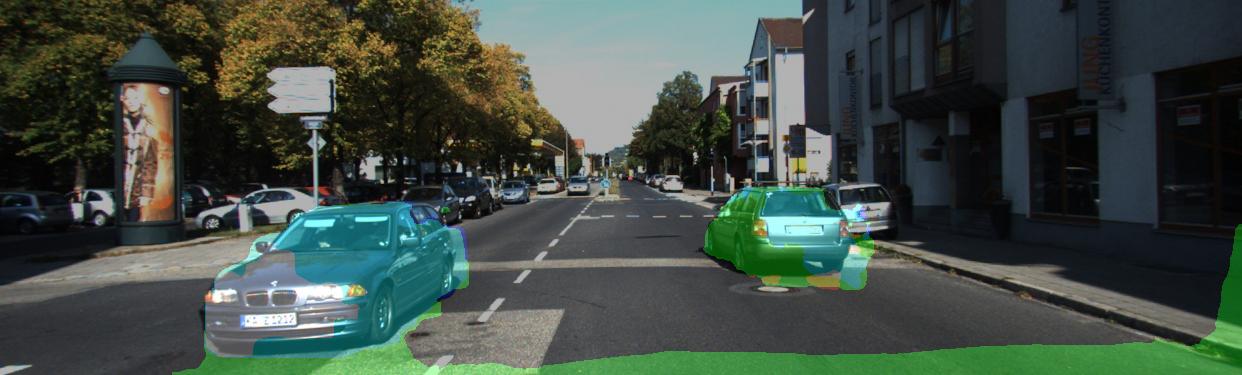} &
\includegraphics[width=0.23\linewidth]{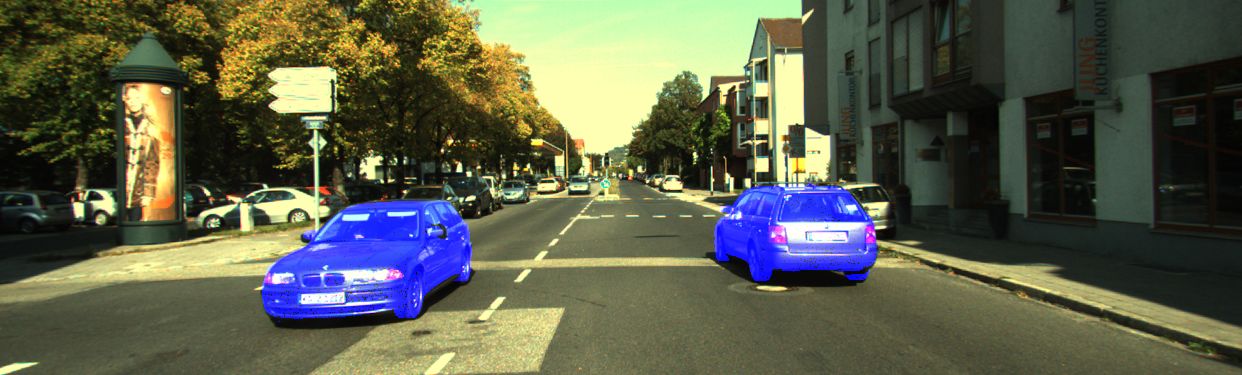} &
\includegraphics[width=0.23\linewidth]{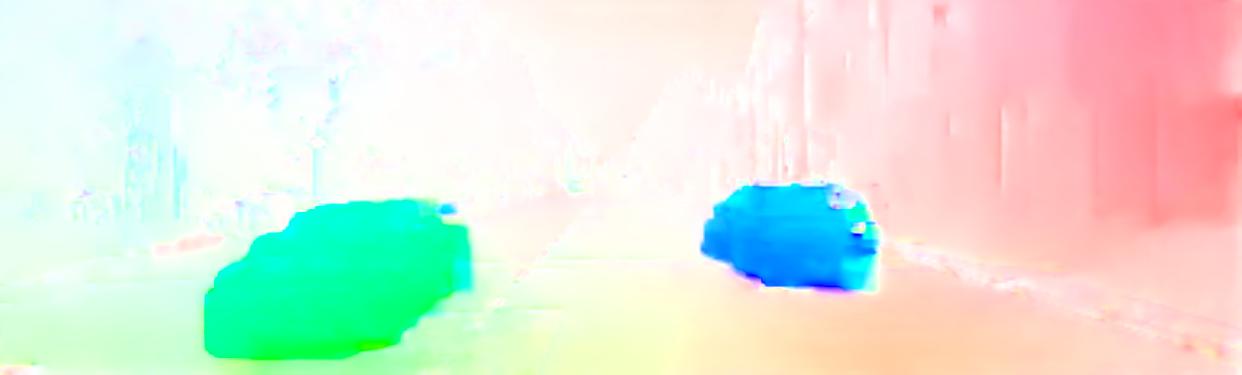} &
\includegraphics[width=0.23\linewidth]{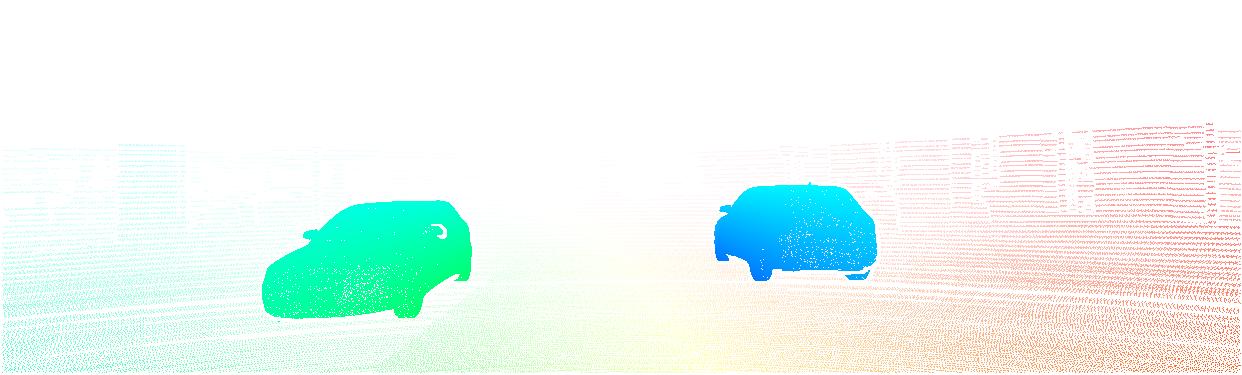} \\
\hline
\hline
\includegraphics[width=0.23\linewidth]{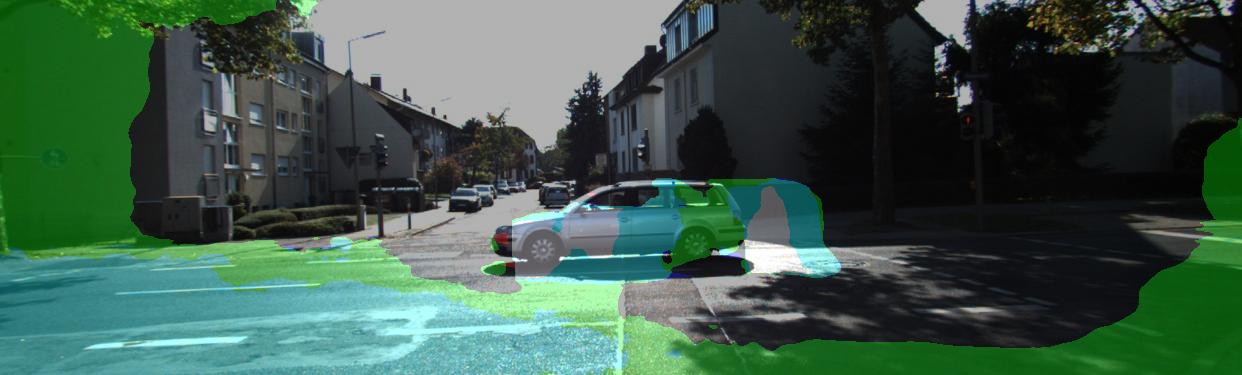} &
\includegraphics[width=0.23\linewidth]{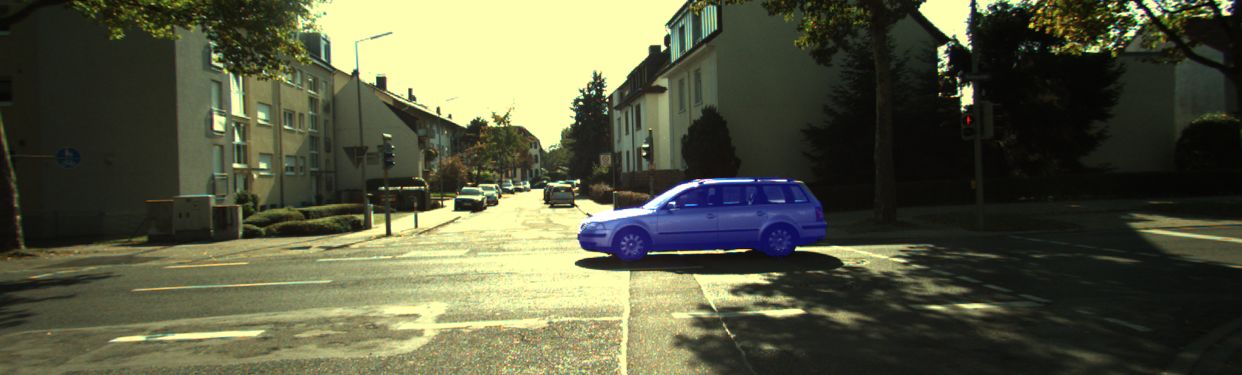} &
\includegraphics[width=0.23\linewidth]{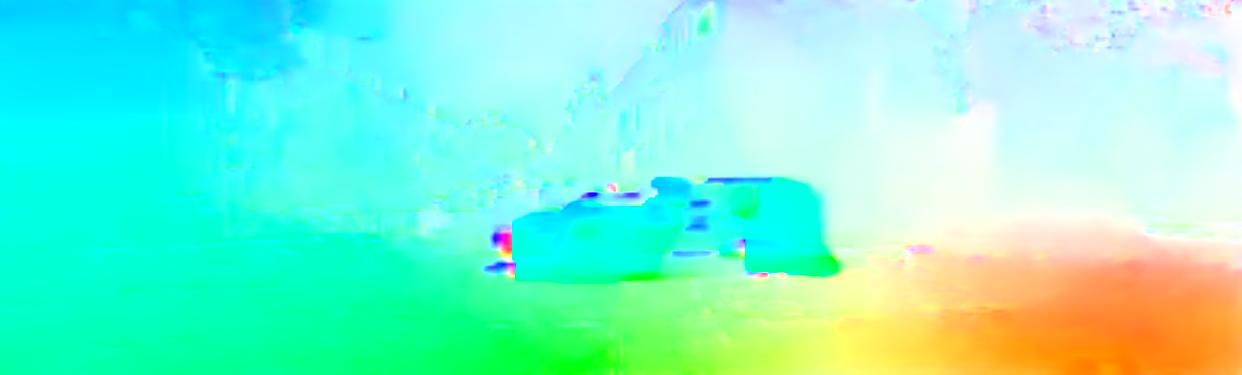} &
\includegraphics[width=0.23\linewidth]{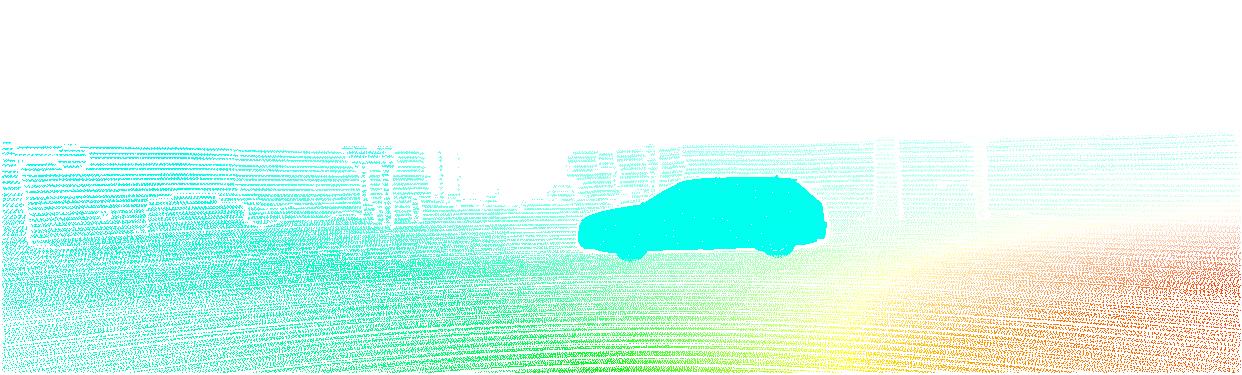} \\
\includegraphics[width=0.23\linewidth]{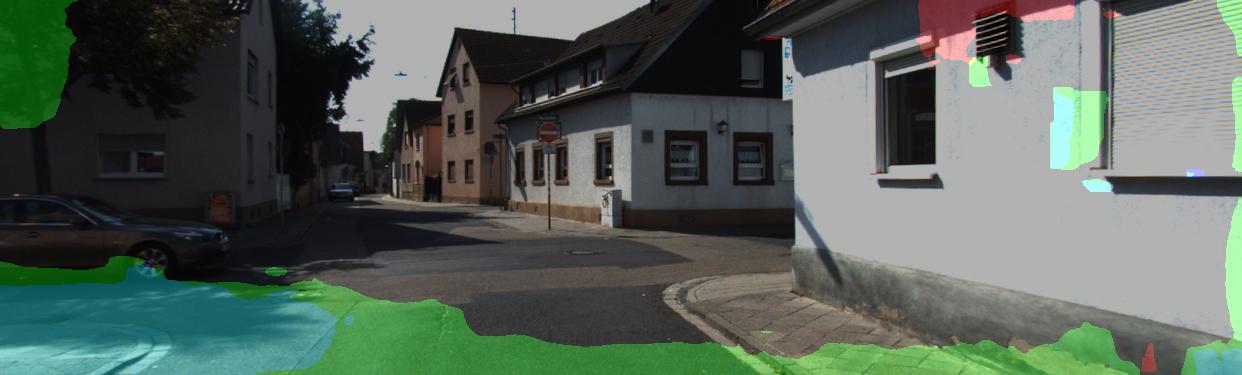} &
\includegraphics[width=0.23\linewidth]{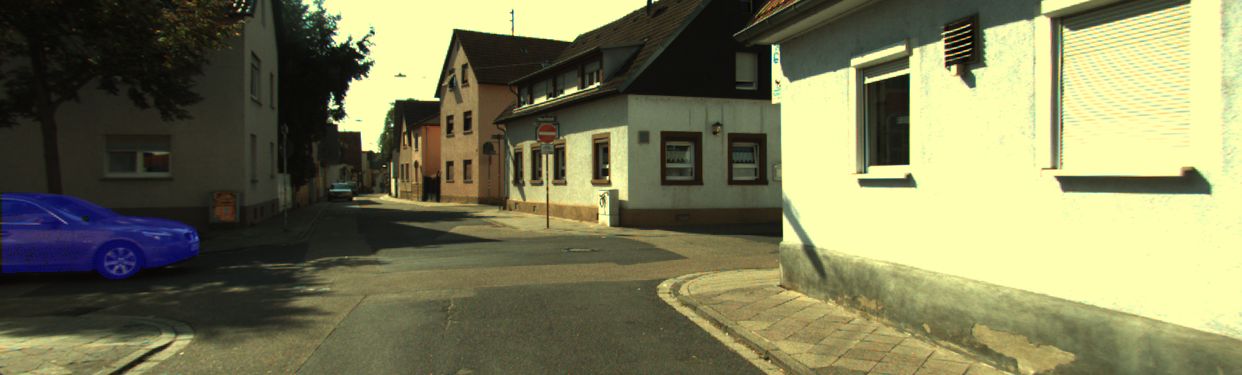} &
\includegraphics[width=0.23\linewidth]{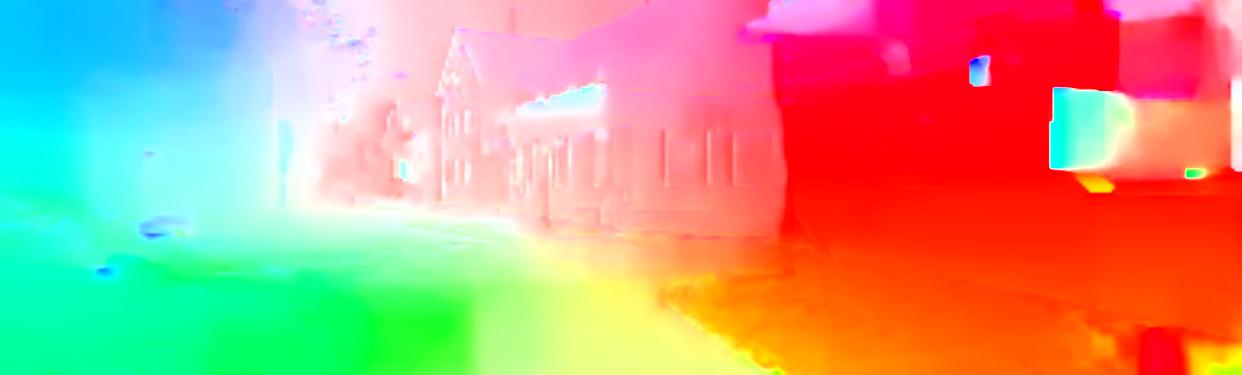} &
\includegraphics[width=0.23\linewidth]{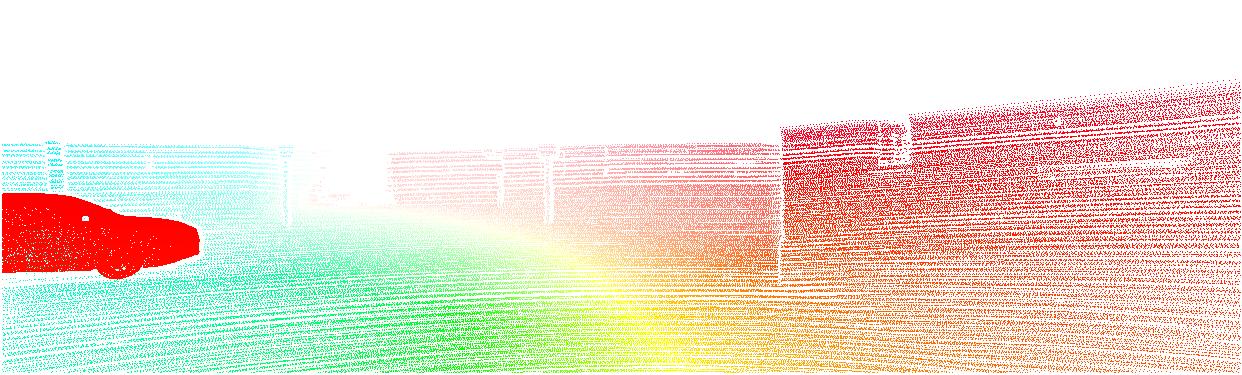} \\

\end{tabular} 
\caption{Ground truth segmentation and flow compared to predicted motion masks and flow from SfM-Net in KITTI 2015. The model was trained in a fully unsupervised manner. The top six rows show successful prediction and the last two show typical failure cases.}
\label{fig:kitti_seg}
\end{figure*}

Figure~\ref{fig:kitti_depth} shows qualitative examples comparing the depth obtained when using stereo pairs with a fixed baseline and when using frame sequences without camera pose information. When there is large translation between the frames, depth estimation without camera pose information is as good as using stereo pairs. The failure cases in the last two rows show that the network did not learn to accurately predict depth for scenes where it saw little or no translation between the frames during training. This is not the case when using stereo pairs as there is always a constant offset between the frames. Using more data could help here because it increases the likelihood of generic scenes appearing in a sequence containing interesting camera motion.

Figure~\ref{fig:kitti_seg} provides qualitative examples of the predicted motion masks and flow fields along with the ground-truth in the KITTI 2015 dataset. Often, the predicted motion masks are fairly close to the ground truth and help explain part of the motion in the scene. We notice that object masks tended to miss very small, distant moving objects. This may be due to the fact that these objects and their motions are too small to be separated from the background. The bottom two rows show cases where the predicted masks do not correspond to moving objects. In the first example, although the mask is not semantically meaningful, note that the estimated flow field is reasonable, with some mistakes in the region occluded by the moving car. In the second failure case, the moving car on the left is completely missed but the motion of the static background is well captured. This is a particularly difficult example for the self-supervised photometric loss because the moving object appears in heavy shadow.

Analysis of our failure cases suggest possible directions for improvement. Moving objects introduce significant occlusions, which should be handled carefully. Because our network has no direct supervision on object masks or object motion, it does not necessarily learn that object and camera motions should be different. These priors could be built into our loss or learned directly if some ground-truth masks or object motions are provided as explicit supervision.

\paragraph{MoSeg.}
The moving objects in KITTI are primarily vehicles, which undergo rigid-body transformations, making it a good match for our model. To verify that our network can still learn in the presence of non-rigid motion, we retrained it from scratch under self-supervision on the MoSeg dataset, using frames from all sequences. Because each motion mask corresponds to a rigid 3D rotation and translation, we do not expect a single motion mask to capture a deformable object. Instead, different rigidly moving object parts will be assigned to different masks. This is not a problem from the perspective of accurate camera motion estimation, where the important issue is distinguishing pixels whose motion is caused by the camera pose transformation directly from those whose motion is affected by independent object motions in the scene.

\begin{figure}
\centering
\begin{tabular}{cccccc}
\textbf{RGB frame} & \textbf{Predicted flow} & \textbf{Motion masks} \\
\includegraphics[width=0.27\linewidth]{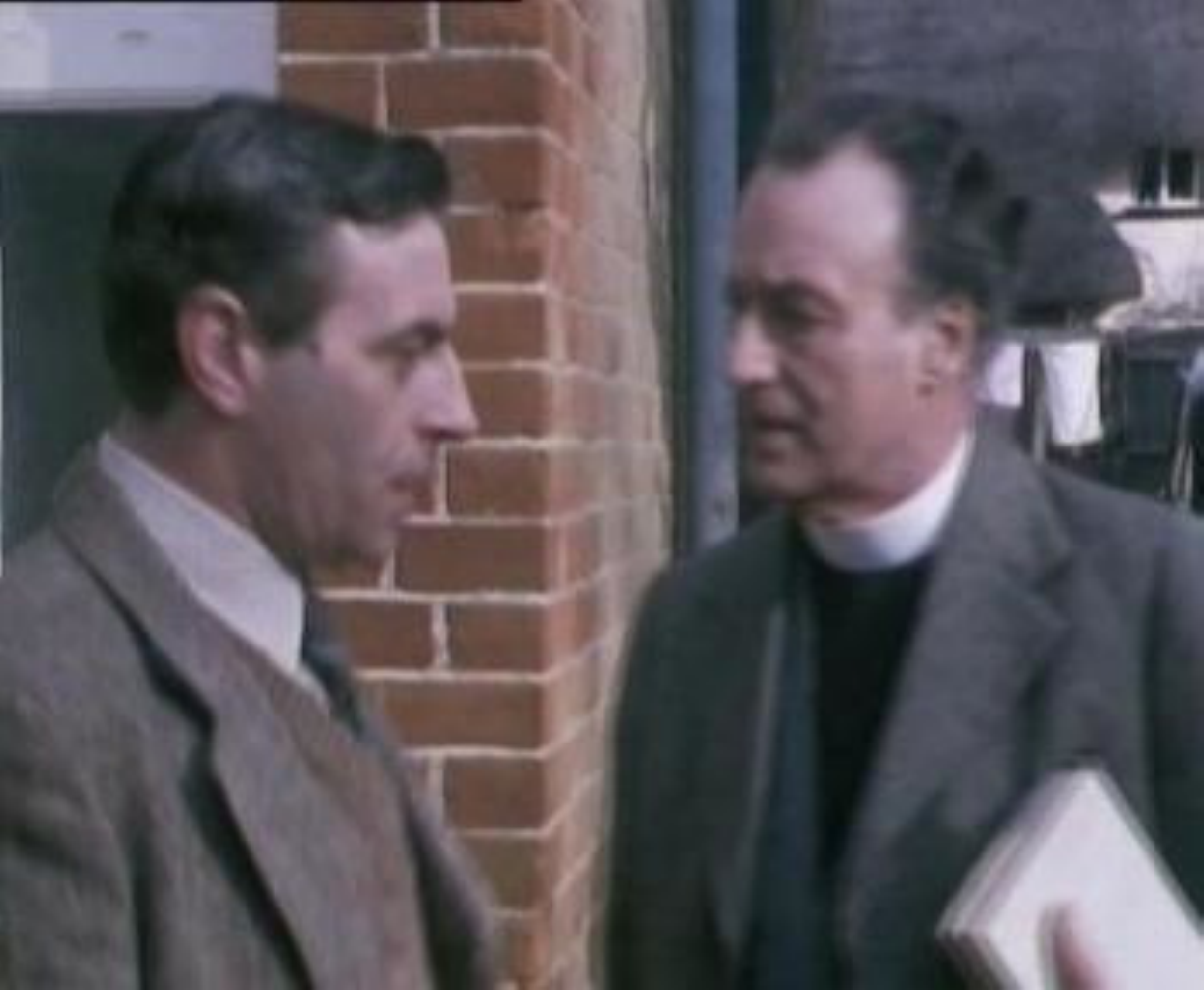} &
\includegraphics[width=0.27\linewidth]{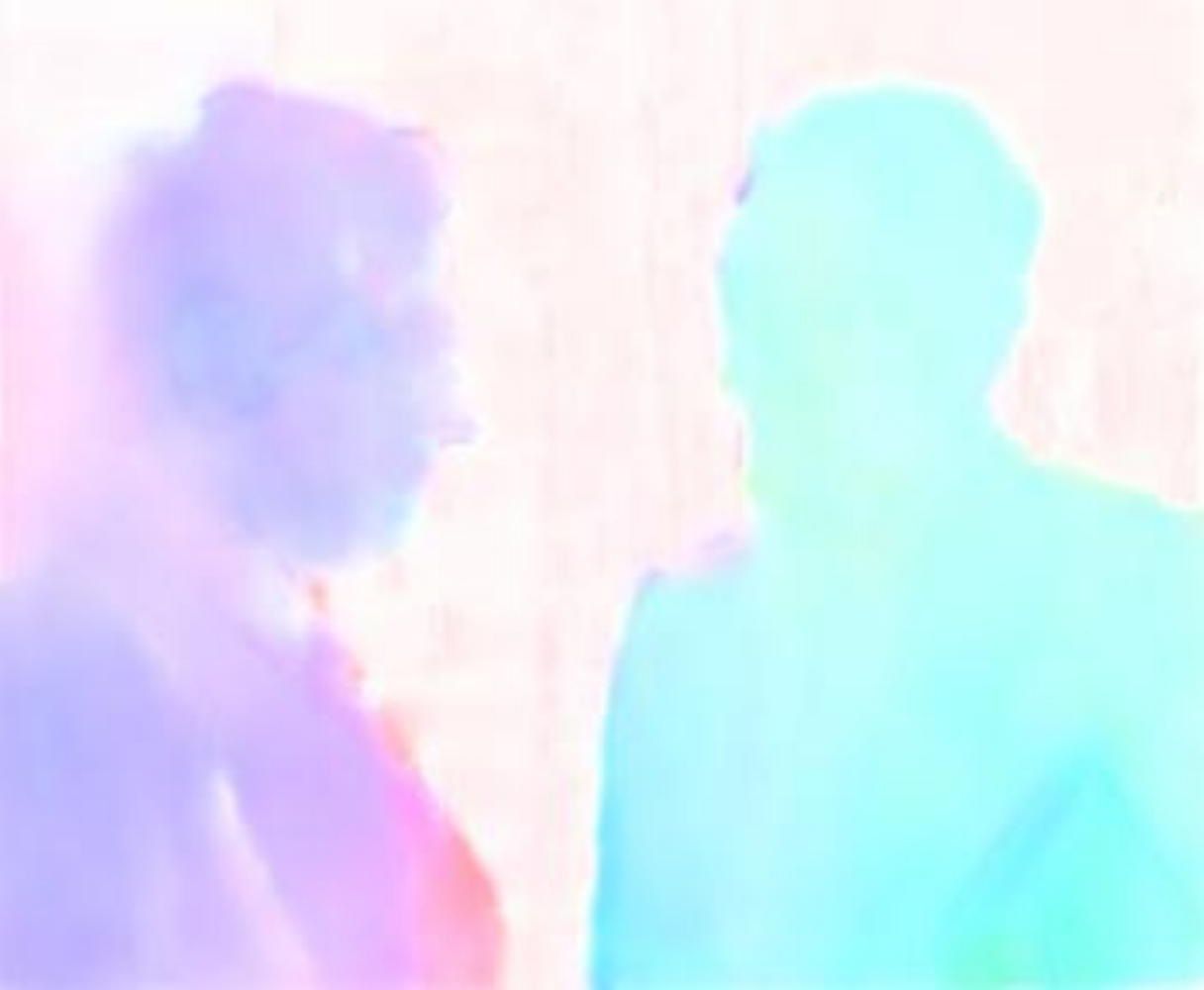} &
\includegraphics[width=0.27\linewidth]{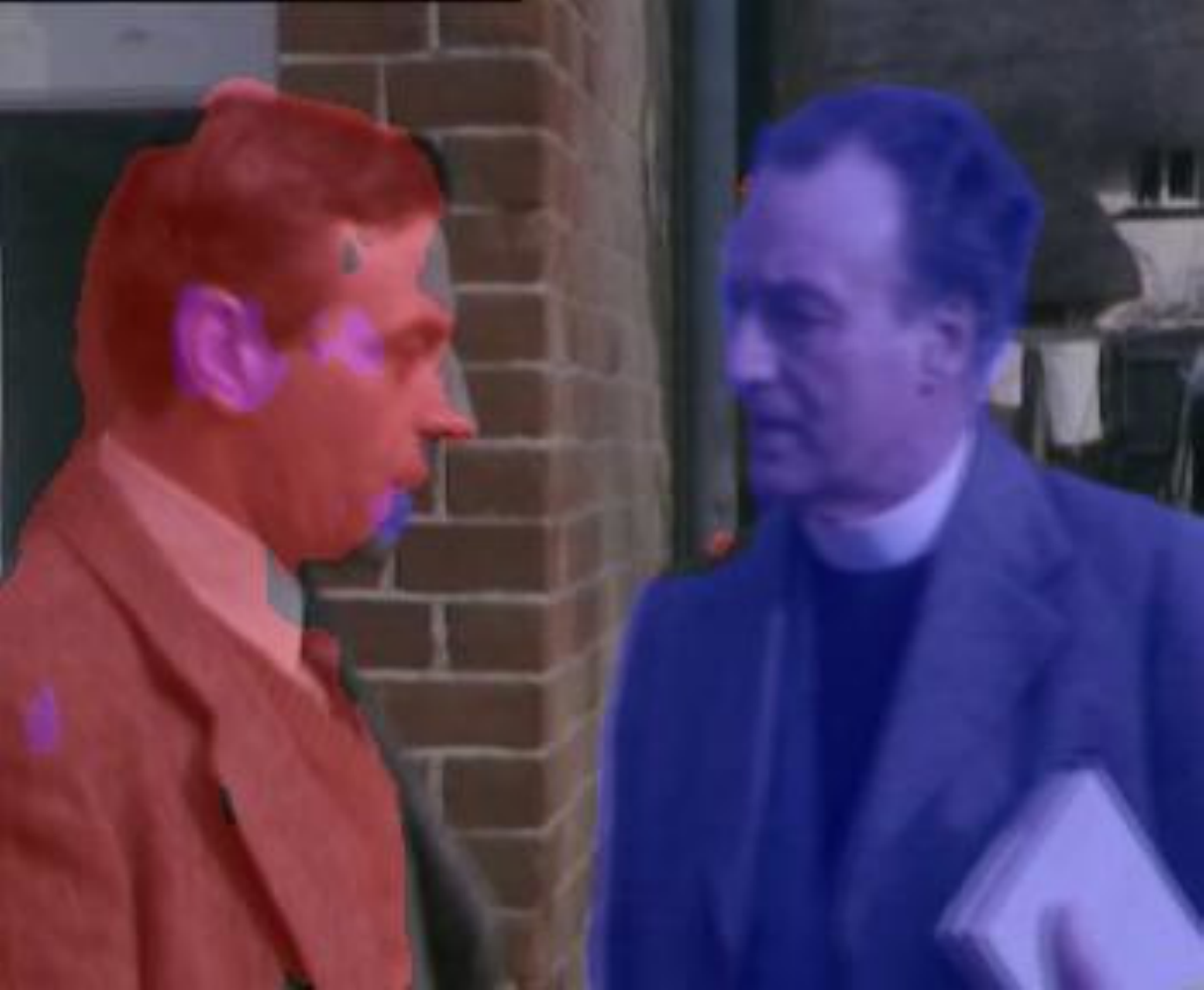} \\
\includegraphics[width=0.27\linewidth]{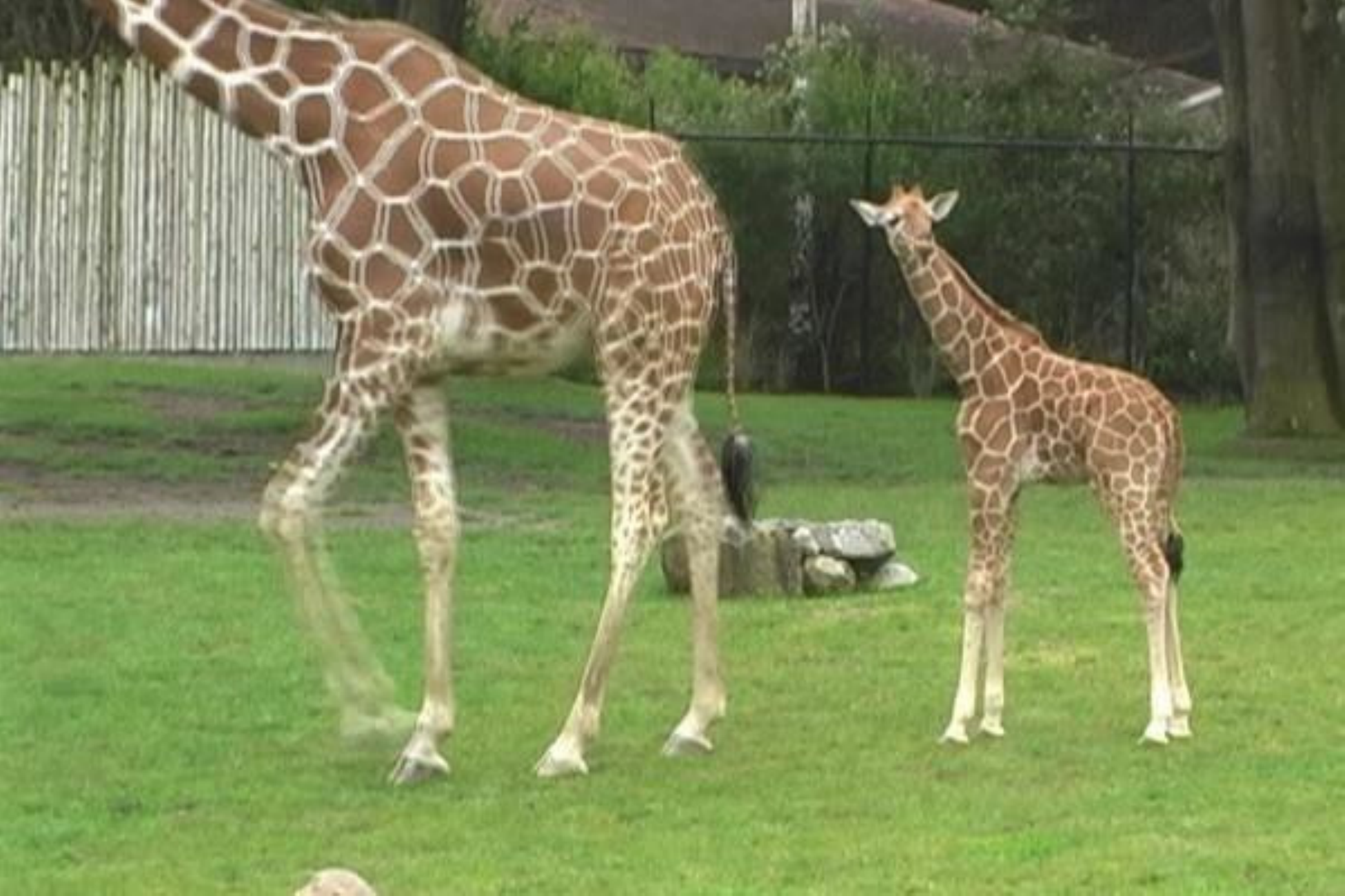} &
\includegraphics[width=0.27\linewidth]{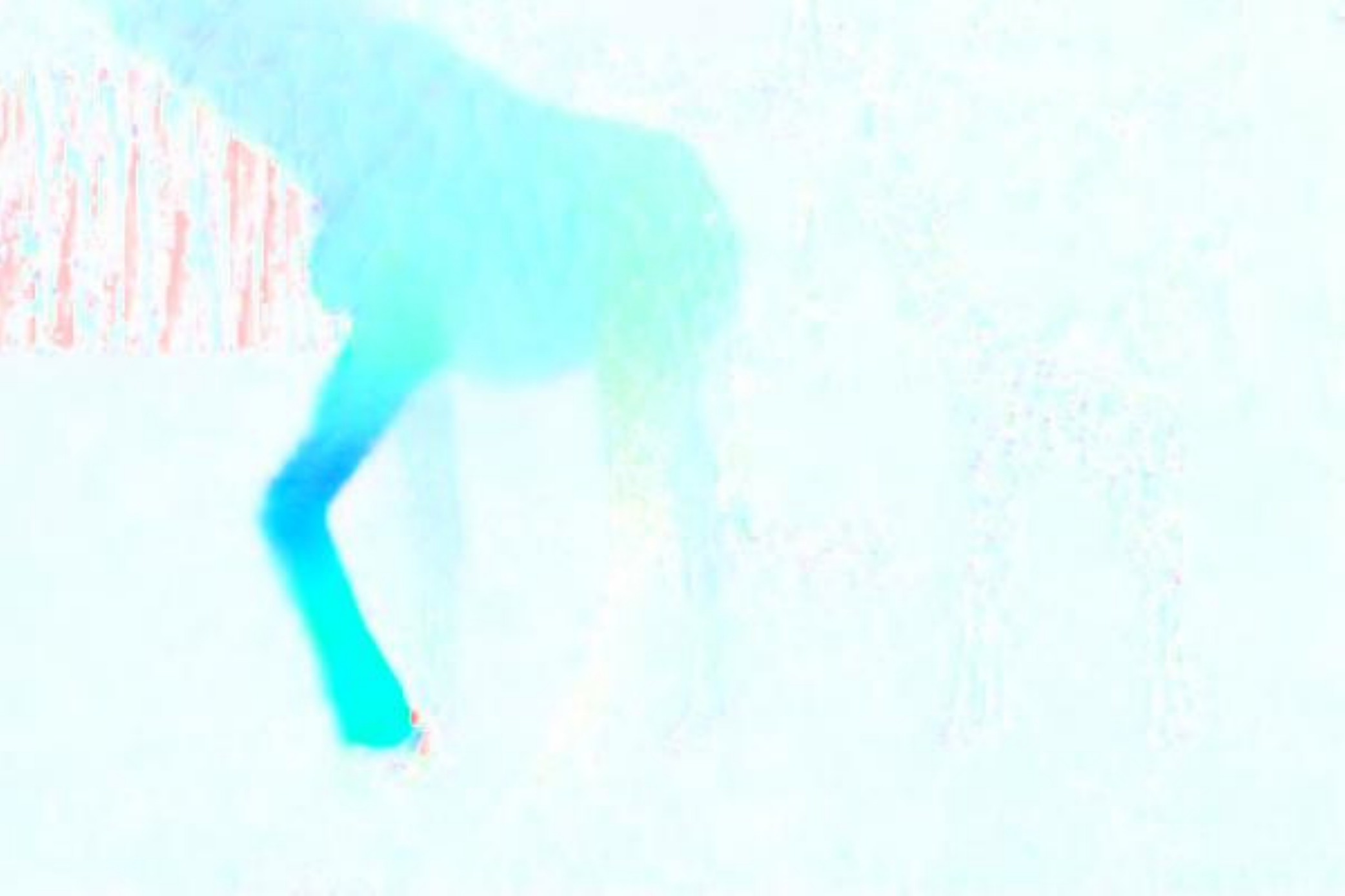} &
\includegraphics[width=0.27\linewidth]{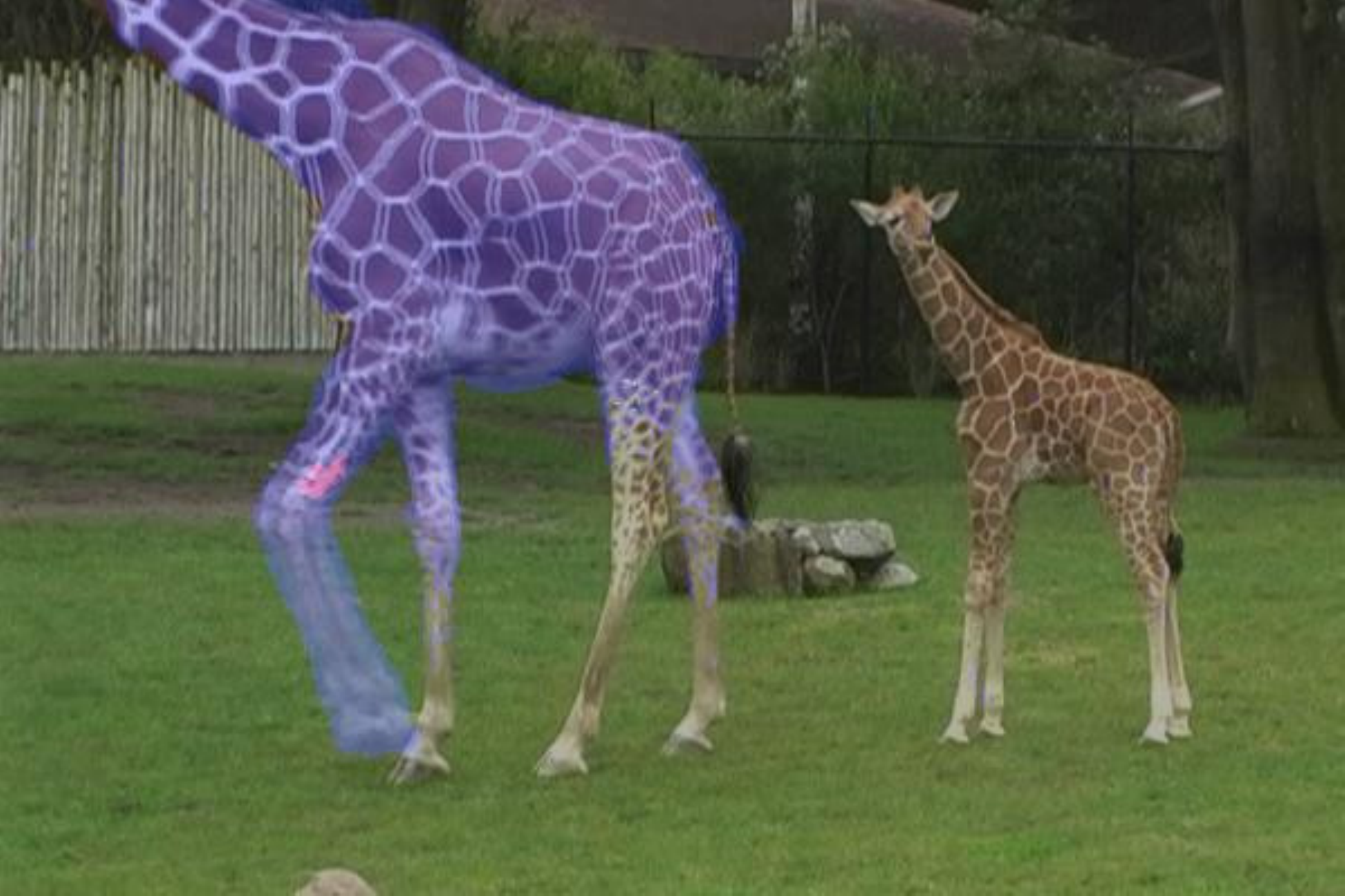} \\
\includegraphics[width=0.27\linewidth]{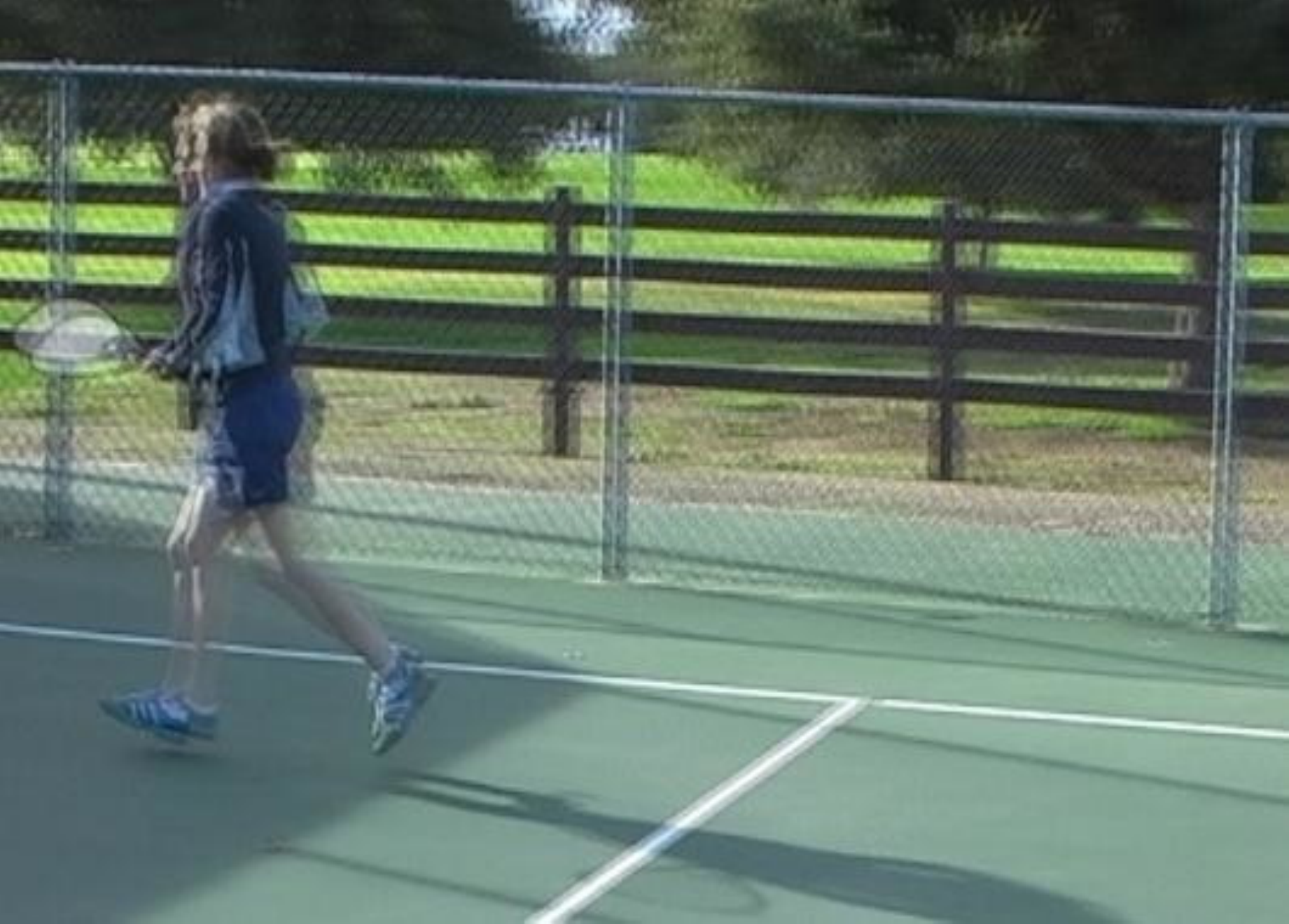} &
\includegraphics[width=0.27\linewidth]{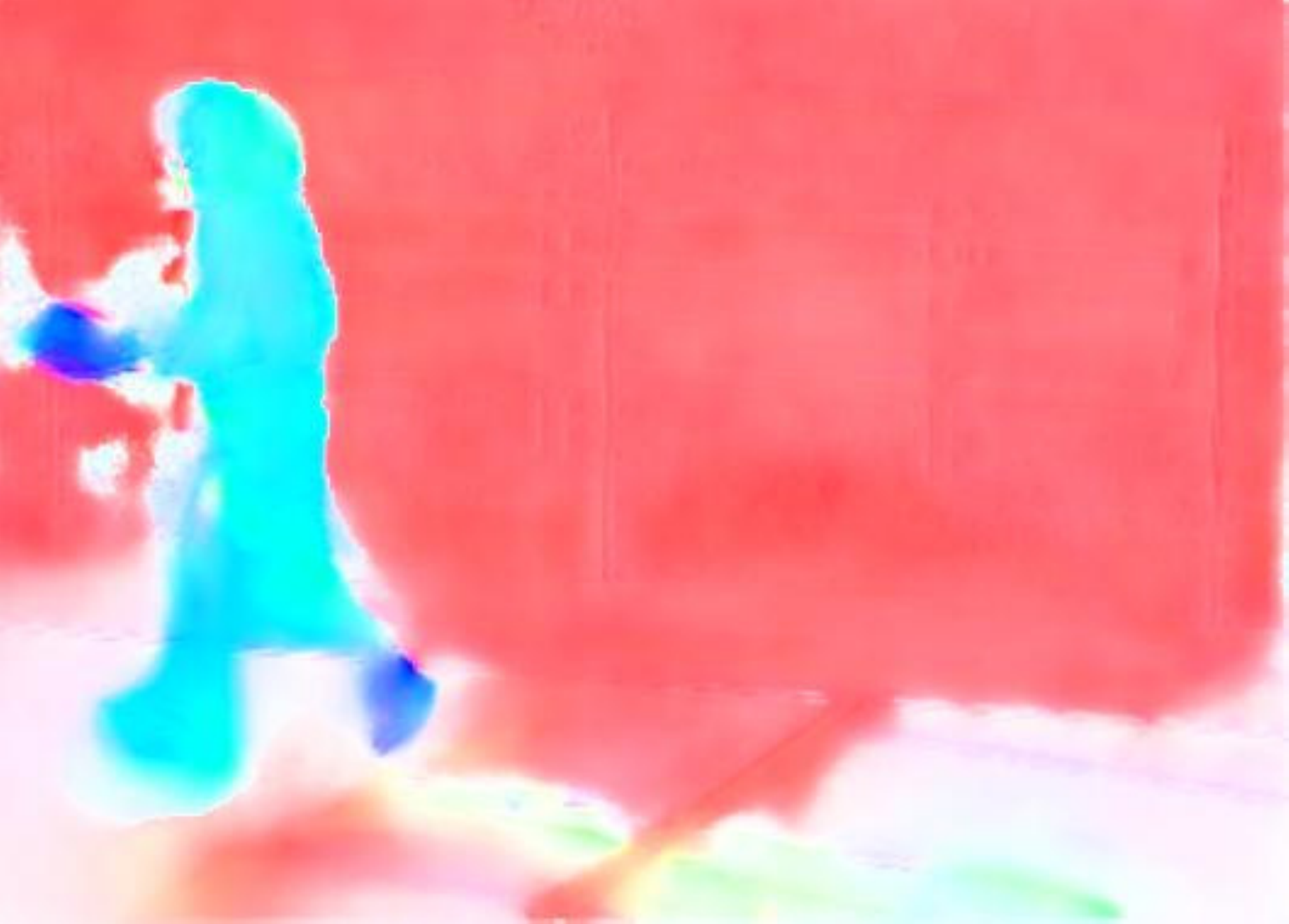} &
\includegraphics[width=0.27\linewidth]{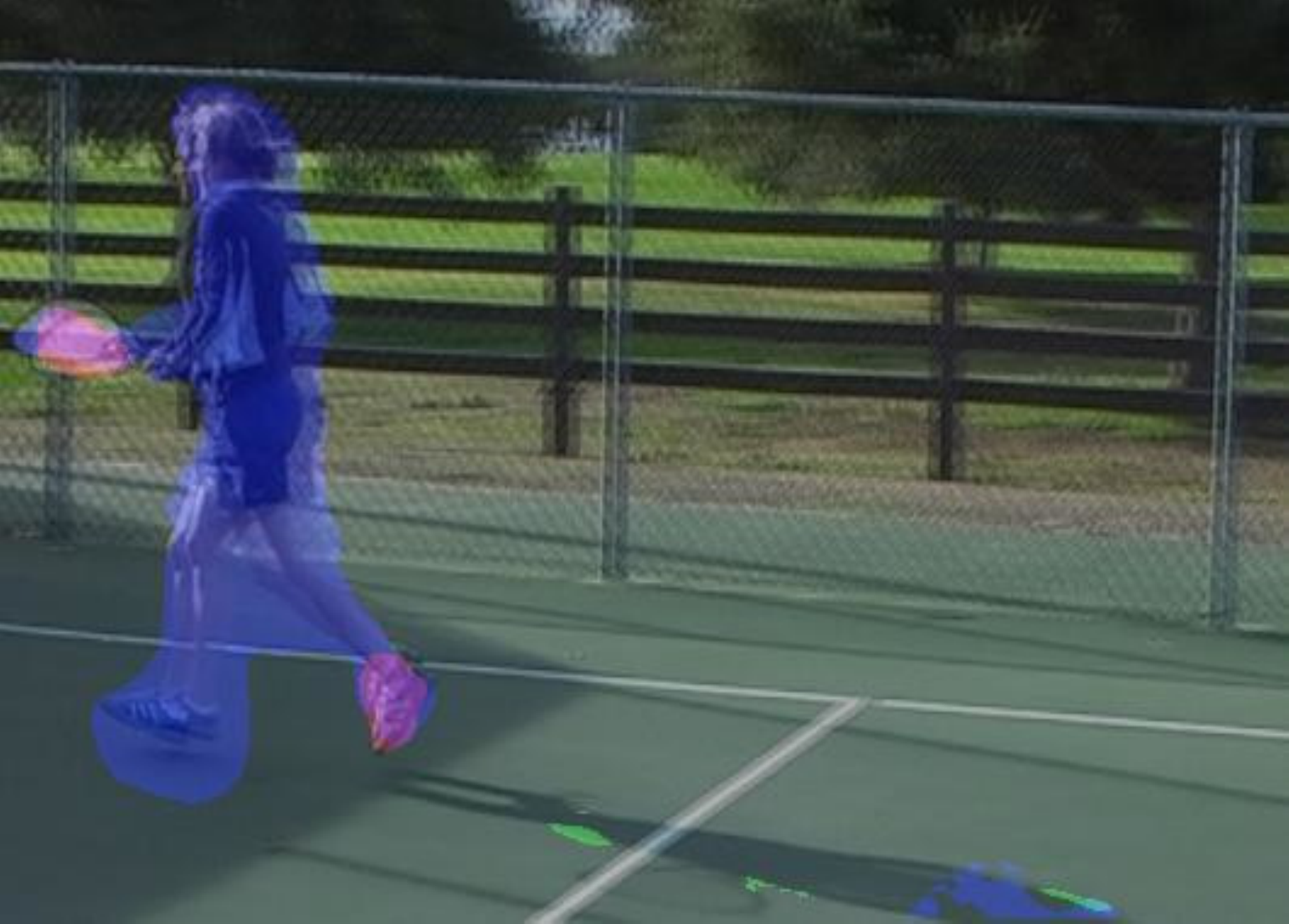} \\
\includegraphics[width=0.27\linewidth]{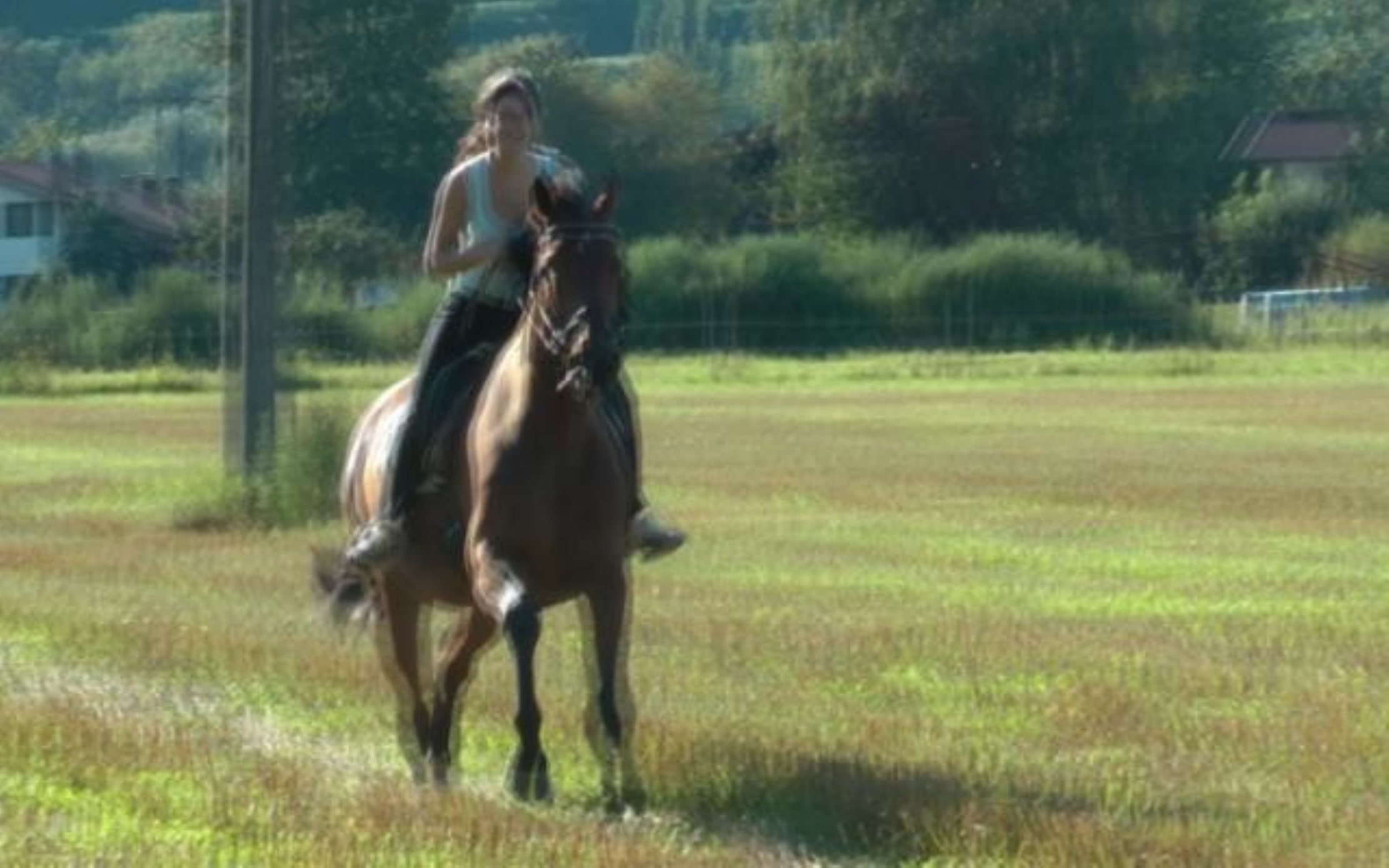} &
\includegraphics[width=0.27\linewidth]{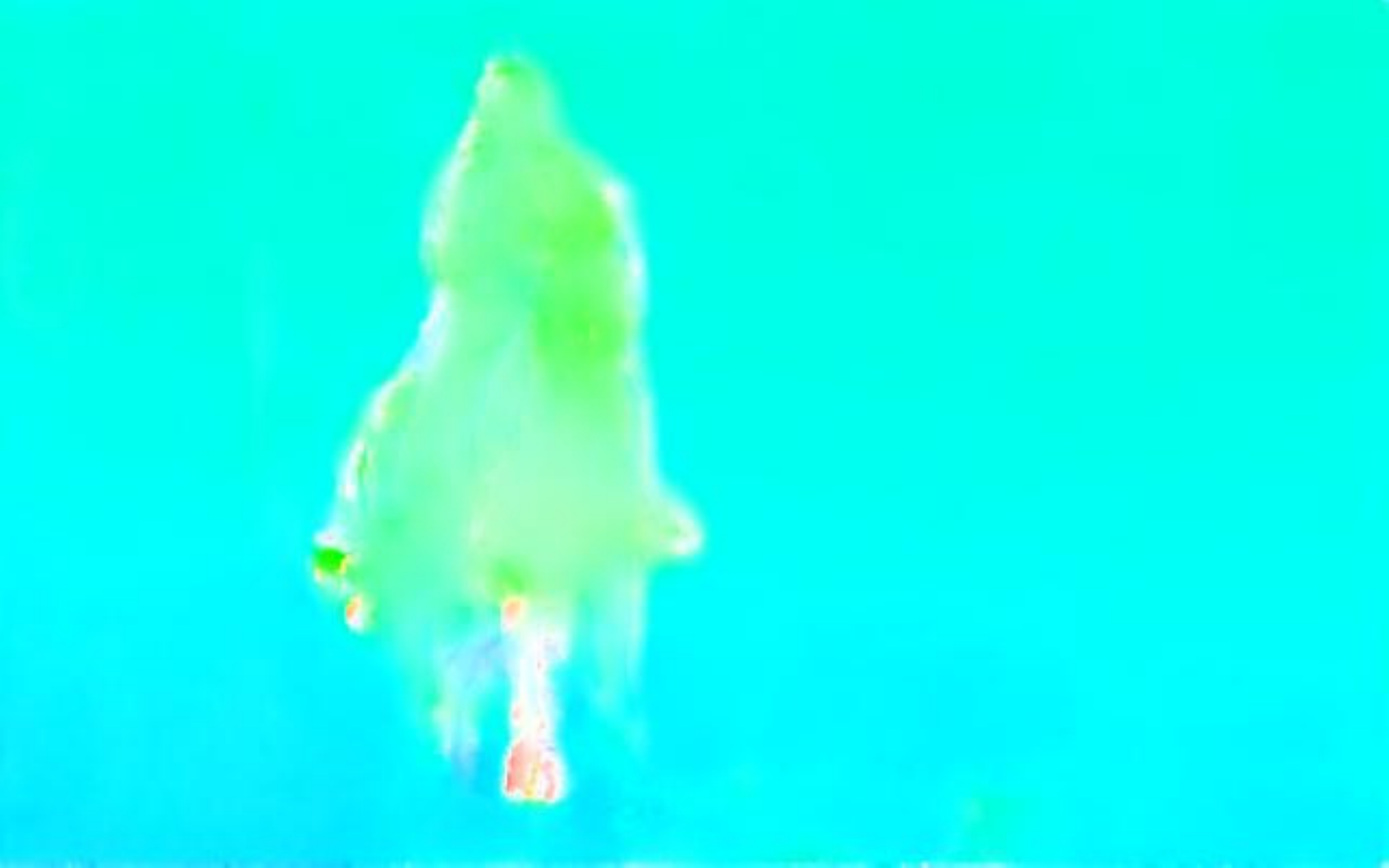} &
\includegraphics[width=0.27\linewidth]{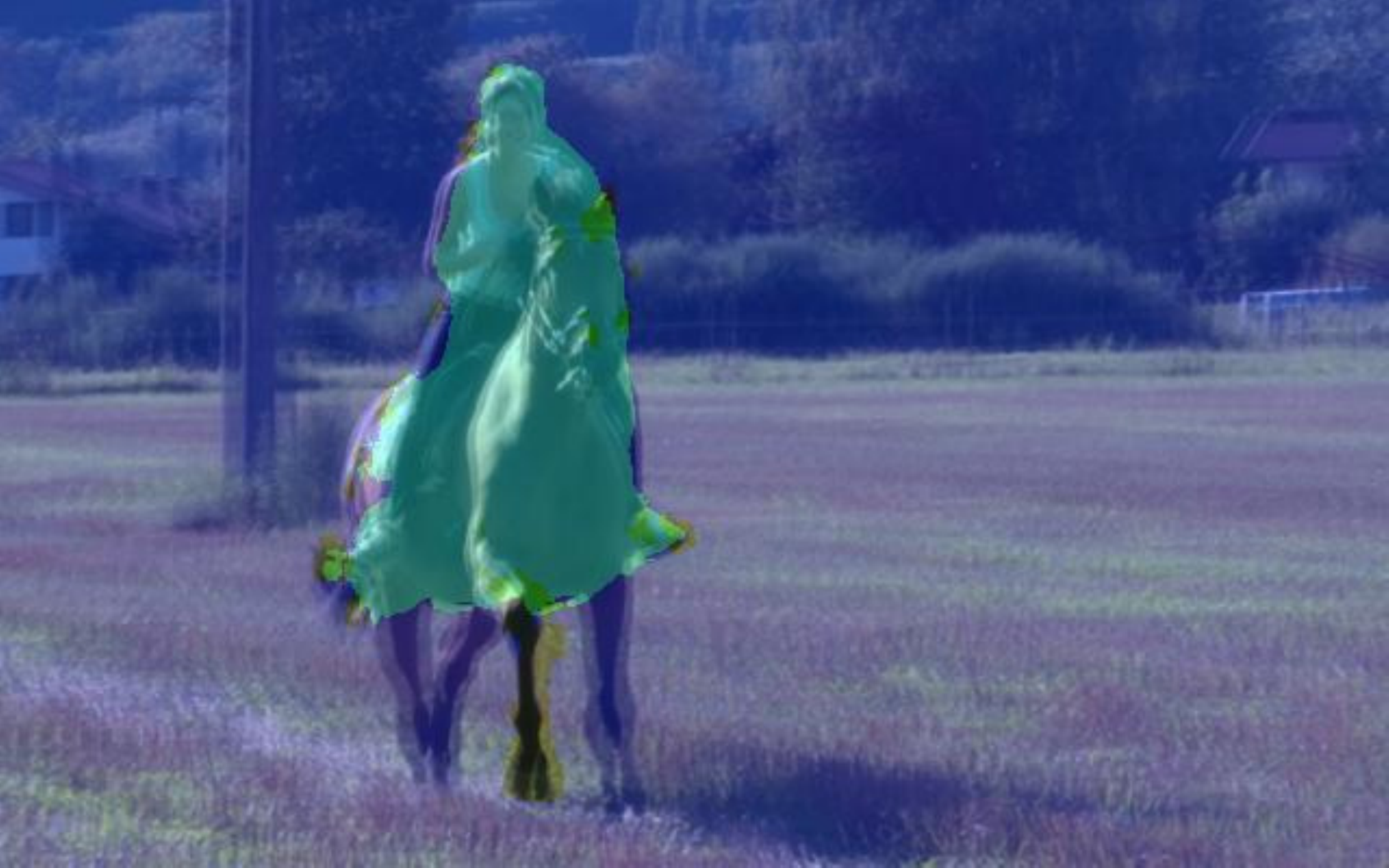} \\
\includegraphics[width=0.27\linewidth]{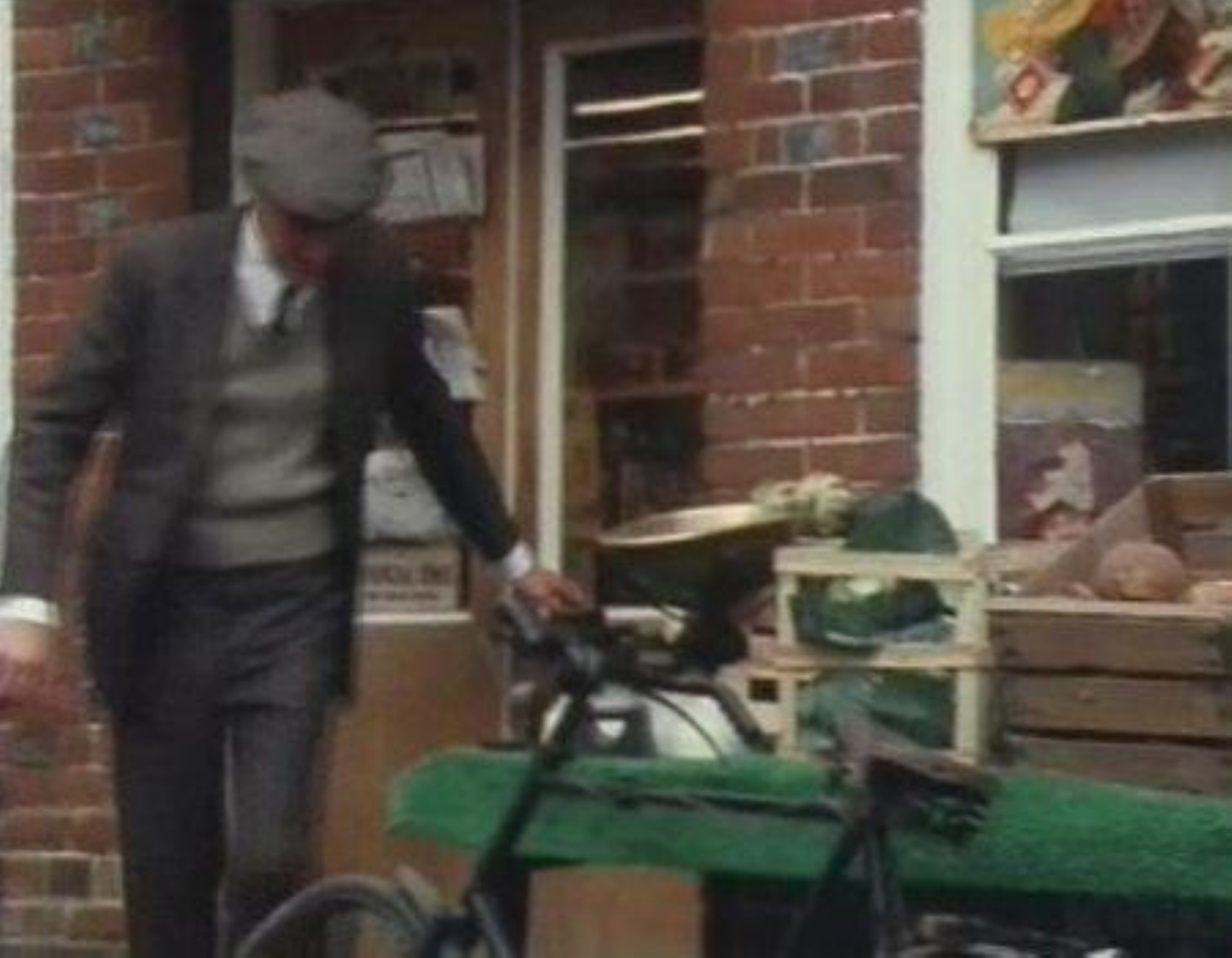} &
\includegraphics[width=0.27\linewidth]{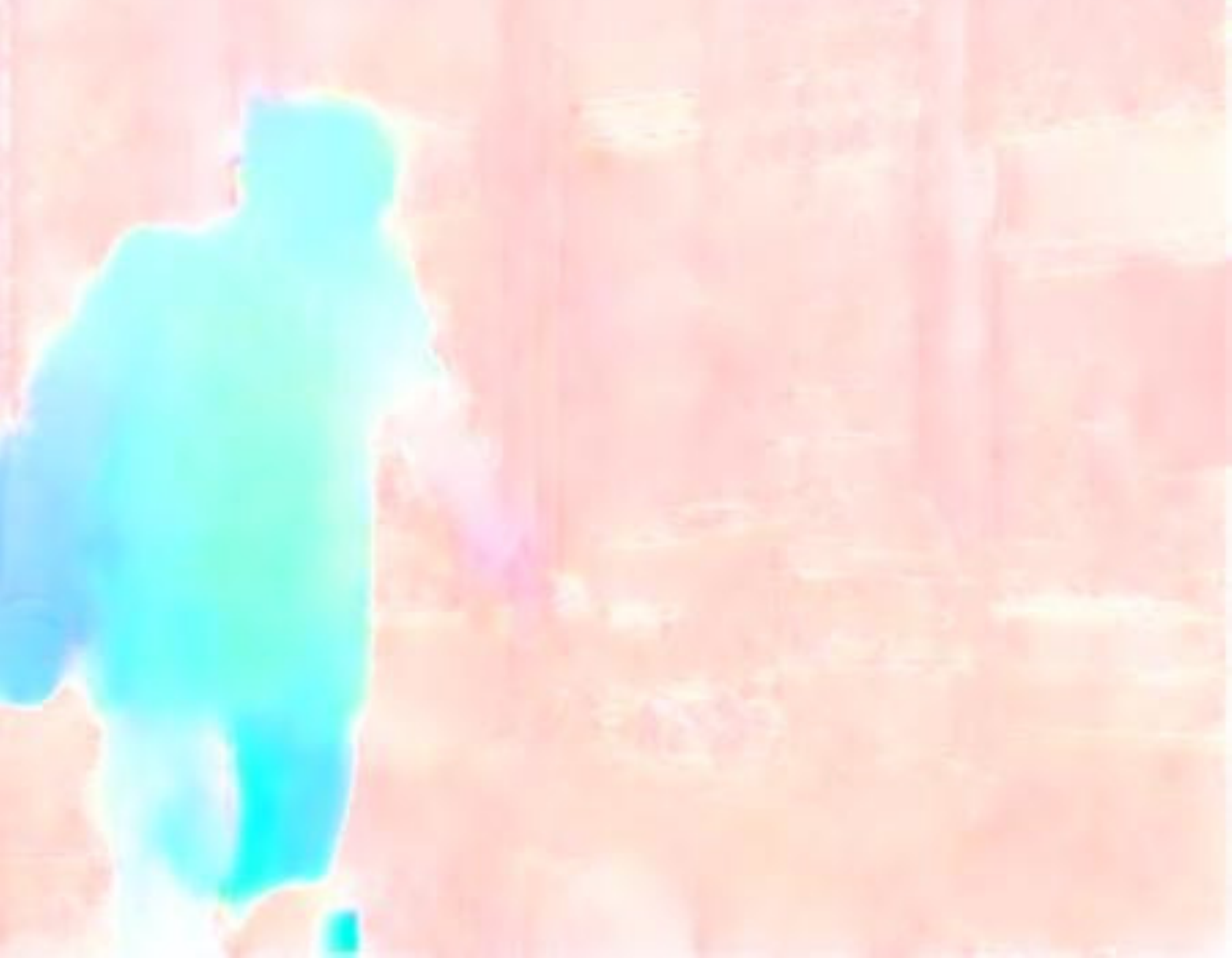} &
\includegraphics[width=0.27\linewidth]{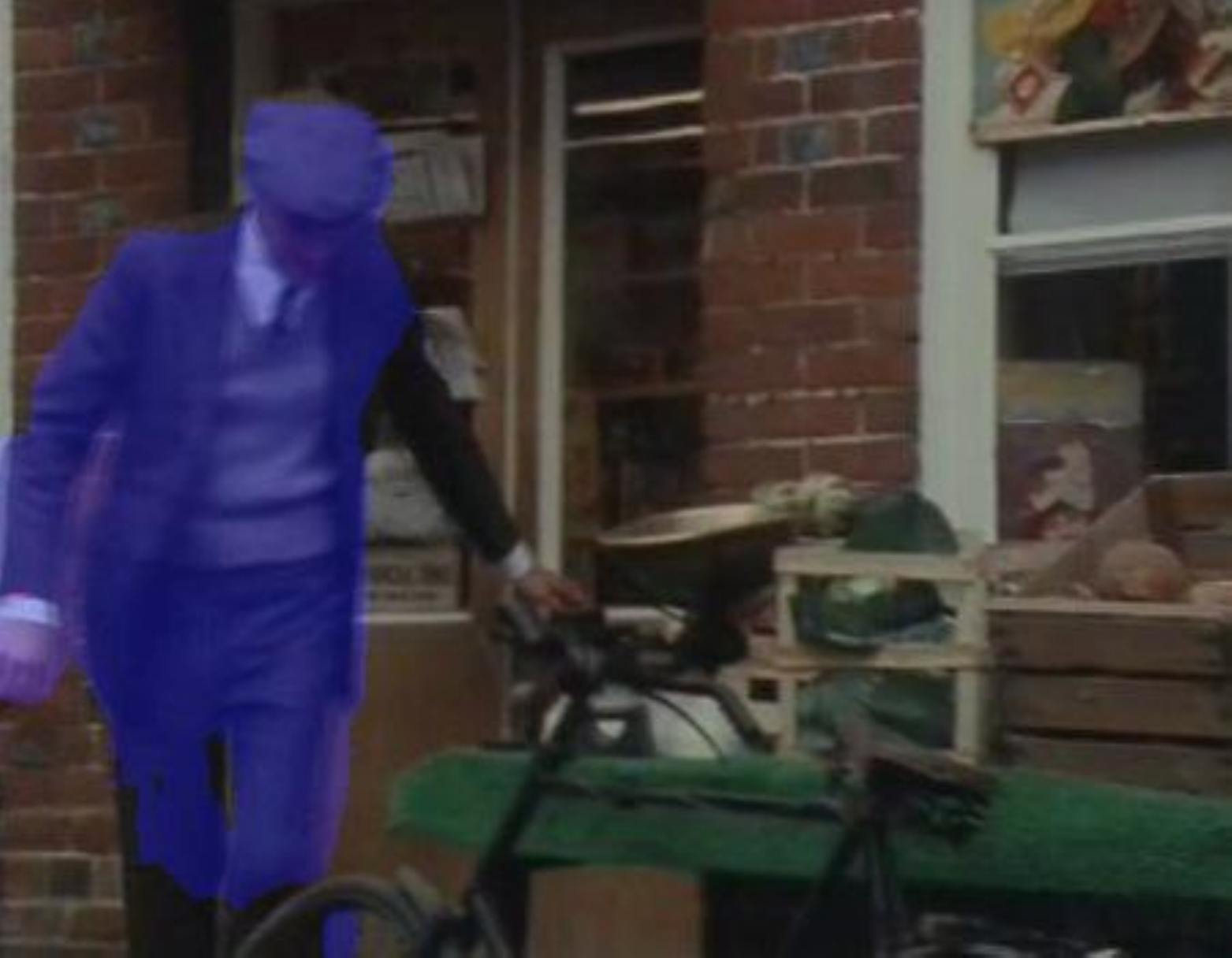} \\
\end{tabular}
\caption{Motion segments computed from SfM-Net in MoSeg \cite{springerlink:10.1007/978-3-642-15555-0_21}. The model was trained in a fully unsupervised manner. 
}
\label{fig:moseg}
\end{figure}

Qualitative results on sampled frames from the dataset are shown in Fig.~\ref{fig:moseg}. Because MoSeg only contains ground-truth annotations for segmentation, we cannot quantitatively evaluate the estimated depth, camera trajectories, or optical flow fields. However, we did evaluate the quality of the object motion masks by computing Intersection over Union (IoU) for each ground-truth segmentation mask against the best matching motion mask and its complement (a total of six proposed segments in each frame, two from each of the three motion masks), averaging across frames and ground-truth objects. We obtain an IoU of 0.29 which is similar to previous unsupervised approaches for the small number of segmentation proposals we use per frame. See, for example, the last column of Figure 5 from  \cite{DBLP:journals/corr/FragkiadakiAFM14}, whose proposed methods for moving object proposals achieve IoU around 0.3 with four proposals. They require more than 800 proposals to reach an IoU above 0.57.

\paragraph{Kinect depth supervision.}

While the fully unsupervised results show promise, our network can benefit from extra supervision of depth or camera motion when available. The improved depth prediction given ground truth camera poses on KITTI stereo demonstrate some gain. We also experimented with adding depth supervision to improve camera motion estimation using the RGB-D SLAM dataset~\cite{sturm12iros}. Given ground-truth camera pose trajectories, we estimated relative camera pose (camera motion) from each frame to the next and compare with the predicted camera motion from our model, by measuring translation and rotation error of their relative transformation, as done in the corresponding evaluation script for relative camera pose error and detailed in Eq.~\ref{eq:cameraposerr}. We report camera rotation and translation error in Table~\ref{rgbdslamcamerapose} for each of the Freiburg1 sequences compared to the error in the benchmark's baseline trajectories. Our model was trained from scratch for each sequence and used the focal length value provided with the dataset. We observe that our results better estimate the frame-to-frame translation and are comparable for rotation. 
\begin{table}
\centering
    \begin{tabular}{|c|c|c|c|c|}
    \hline
         Seq. & transl \cite{sturm12iros} & rot \cite{sturm12iros} & transl. ours & rot ours \\
         \hline
         360 & 0.099    & 0.474& 0.009 &  1.123\\
         plant & 0.016& 1.053& 0.011 &   0.796 \\
         teddy & 0.020 &  1.14 & 0.0123&  0.877 \\
         desk &  0.008& 0.495&  0.012&   0.848 \\
         desk2 & 0.099 &  0.61& 0.012&  0.974  \\ 
         \hline 
        
    \end{tabular}
    \caption{Camera pose relative error from frame to frame for various video sequences of Freiburg RGBD-SLAM benchmark.}
    \label{rgbdslamcamerapose}
\end{table}

\section{Conclusion}
\label{sec:conclusion}

Current geometric SLAM methods obtain excellent ego-motion and rigid 3D reconstruction results, but often come at a price of extensive engineering, low tolerance to moving objects --- which are treated as noise during reconstruction --- and sensitivity to camera calibration. Furthermore, matching and reconstruction are difficult in low textured regions.  
Incorporating learning into depth reconstruction, camera motion prediction and object segmentation, while still preserving the constraints of image formation, is a promising way to robustify SLAM and visual odometry even further. However, the exact training scenario required to solve this more difficult inference problem remains an open question. Exploiting long history and far in time forward-backward constraints with visibility reasoning is  an important future direction. Further, exploiting a small amount of annotated videos for object segmentation, depth, and camera motion,  and combining those with an abundance of self-supervised videos, could help initialize the network weights in the right regime and facilitate learning. Many other curriculum learning regimes, including those that incorporate synthetic datasets, can also be considered.

\paragraph{Acknowledgements.} We thank our colleagues Tinghui Zhou, Matthew Brown, Noah Snavely, and David Lowe for their advice and Bryan Seybold for his work generating synthetic datasets for our initial experiments.

{\small
\bibliographystyle{ieee}
\bibliography{egbib}
}

\end{document}